\documentclass[10pt,twocolumn,letterpaper]{article}

\usepackage{wacv}              

\definecolor{wacvblue}{rgb}{0.21,0.49,0.74}
\usepackage[pagebackref,breaklinks,colorlinks,allcolors=wacvblue]{hyperref}

\usepackage{bm}
\usepackage{multirow}
\usepackage{adjustbox}
\usepackage{makecell}

\usepackage{graphicx}
\usepackage{xcolor}
\usepackage{booktabs}
\usepackage{multirow}
\usepackage{array}
\usepackage{colortbl}
\usepackage{url}
\usepackage{tabularx}
\usepackage{wrapfig}
\usepackage{threeparttable}
\usepackage{algorithm}
\usepackage{algpseudocode}

\newcommand{\proposed}{Stitch4D}

\newcommand{\moduleone}{\textsc{MVBM}\,}

\newcommand{\moduletwo}{\textsc{MVJOM}\,}

\newcommand{\para}[1]{
  \vspace{1.2mm}
  \noindent\textbf{#1. }
}

\usepackage{graphicx}
\usepackage{booktabs}
\usepackage{bm}
\usepackage{multirow}
\usepackage{adjustbox}
\usepackage{makecell}
\usepackage[most]{tcolorbox}
\usepackage{graphicx}
\usepackage{wrapfig}
\usepackage{cuted}
\usepackage{hyperref}
\usepackage{orcidlink}
\usepackage{tablefootnote}

\newcommand{\TColorBox}[2]{
\begin{tcolorbox}[enhanced, breakable,
  colback=blue!5!white,
  colframe=blue!75!black,
  fonttitle=\bfseries,
  title=#1, 
  boxrule=0.5pt,
  sharp corners,
  left=2mm, right=2mm, top=1mm, bottom=1mm,
  before upper={\let\\\par},
]
#2
\end{tcolorbox}
}

\def\wacvPaperID{551} 
\def\confName{WACV}
\def\confYear{2027}

\title{Stitch4D: 4D Urban Reconstruction from Sparse Multi-Location Observations via Spatio-Temporal Interpolation}

\author{
Hina Kogure\textsuperscript{1,*} \quad
Kei Katsumata\textsuperscript{1,*} \quad
Taiki Miyanishi\textsuperscript{2,1} \quad
Komei Sugiura\textsuperscript{1}\\
\textsuperscript{1}Keio University, Japan \quad
\textsuperscript{2}The University of Tokyo, Japan\\
{\tt\small \{hina.chick1717, ke59ka77, komei.sugiura\}@keio.jp}\\
{\tt\small taiki.miyanishi@weblab.t.u-tokyo.ac.jp}
}

\begin{document}
\maketitle

\begingroup
\renewcommand{\thefootnote}{\fnsymbol{footnote}}
\footnotetext[1]{Equal contribution.}
\endgroup

\begin{abstract}
Dynamic urban environments are often captured by cameras placed at spatially separated locations with little or no view overlap.
However, most existing 4D reconstruction methods assume densely overlapping views and struggle under sparse multi-location observations, producing unstable reconstructions in unobserved intermediate regions.
To address this practical yet underexplored setting, we propose \textbf{Stitch4D}, a unified 4D reconstruction framework that compensates for missing spatial coverage in sparsely observed urban environments.
Stitch4D synthesizes intermediate \emph{bridge views} between distant camera locations and jointly optimizes real and synthesized observations in a unified coordinate frame with inter-location consistency constraints.
By recovering intermediate spatial coverage before optimization, Stitch4D mitigates geometric collapse and improves reconstruction stability in sparse regions.
To evaluate this setting, we introduce \textbf{Urban Sparse 4D (U-S4D)}, a controlled CARLA-based benchmark for free-viewpoint reconstruction under sparse multi-location configurations.
Experiments on U-S4D show that Stitch4D consistently outperforms representative 4D reconstruction baselines in image-quality metrics.
These results suggest that recovering intermediate spatial coverage is an effective strategy for stabilizing 4D reconstruction in sparse urban environments.
The project page is provided in \url{https://stitch4d-project-page.vercel.app/}.
\end{abstract}    
\begin{figure*}[t]
    \vspace{-4mm}
    \centering
    \includegraphics[width=0.85\linewidth]{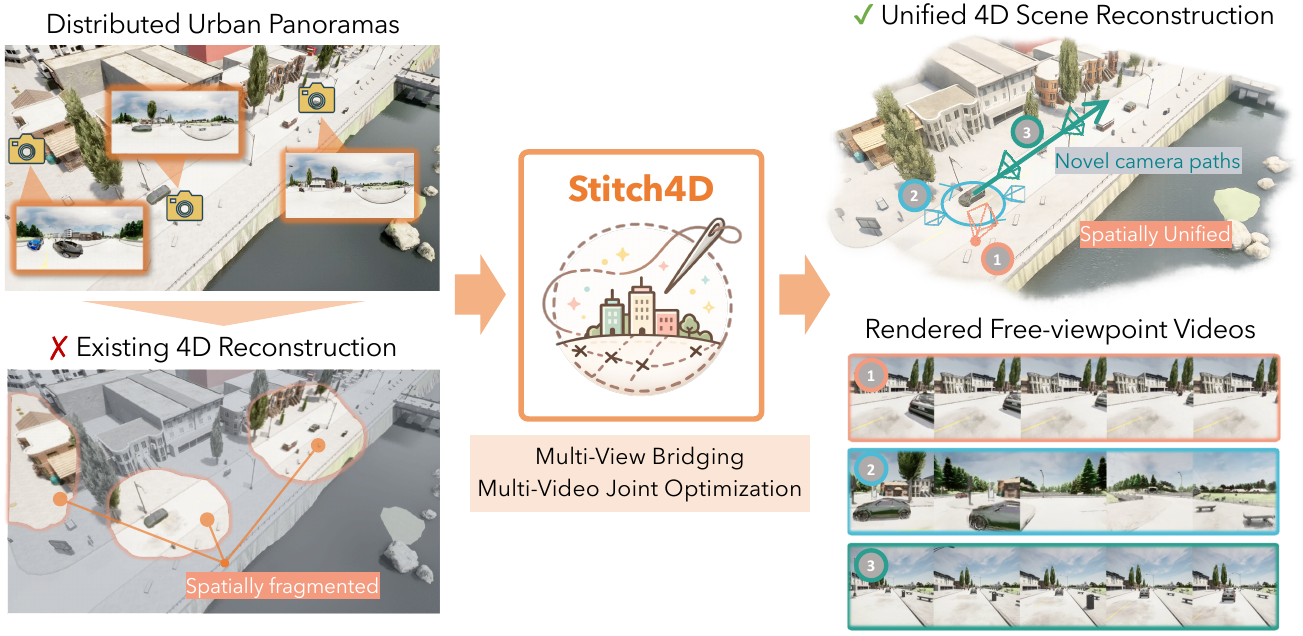}
    \vspace{-2mm}
    \caption{
\textbf{Sparse multi-location 4D reconstruction with Stitch4D.}
%
Existing methods produce spatially fragmented reconstructions from distributed urban panoramas, while Stitch4D bridges missing regions to recover a unified 4D scene for novel-view rendering.
    %
    }
    \vspace{-6mm}
    \label{fig:eye-catch}
\end{figure*}

\vspace{-6mm}
\section{Introduction}
\label{sec:intro}
\vspace{-3mm}
Accurate modeling of dynamic urban environments is essential for autonomous driving, robotics, and urban analytics~\cite{katsumata2025gennav, xie2025compositional, yasuki2025geoprog3d}.
Urban dynamics span multiple spatial and temporal scales, from city-level layouts and traffic flows to the fine-grained motion of individual agents.
These characteristics motivate 4D representations that jointly model scene geometry and dynamics over time~\cite{cao2025reconstructing, kong20253d, liao2025learning}.

%
In real urban deployments, however, camera placement is constrained by infrastructure availability and privacy regulations~\cite{kansal2024implications}.
As a result, observations are often captured from spatially distributed locations with limited overlap in their fields of view.
Such configurations commonly arise in urban surveillance systems with geographically distributed cameras and in fleet-collected driving videos from vehicle-mounted cameras.
Under such sparse multi-location conditions, the core challenge is not merely the sparsity of views, but the lack of reliable overlap across spatially separated observations.
Existing 4D reconstruction methods typically assume sufficient viewpoint overlap and perform well under closely spaced camera setups~\cite{wu20244dgs, li2024spacetime, wang2025freetimegs}.
When applied to images captured by cameras that are far apart, correspondence estimation and dynamic tracking become unreliable, leading to geometric misalignment and temporal inconsistencies~\cite{han2025extrapolated}.


\vspace{-1mm}
To study this practical setting, we formulate the Sparse Multi-Location 4D Reconstruction (SP4DR) problem, which reconstructs a unified spatiotemporal scene representation from time-synchronized videos captured at spatially separated locations.
SP4DR requires jointly modeling dynamic entities, such as vehicles and pedestrians, together with static structures while preserving geometric consistency across locations.
Unlike conventional multi-view capture setups, cameras in SP4DR may have little or no viewpoint overlap, making correspondence estimation and spatiotemporal alignment more challenging.

To address this challenge, we propose \proposed, a unified framework for SP4DR that bridges spatial gaps between distant camera locations and jointly optimizes observations across locations.
Fig.~\ref{fig:eye-catch} illustrates the SP4DR setting and the overview of \proposed, which stitches spatially separated observations into a unified 4D representation.
Rather than reconstructing each camera location independently, \proposed~explicitly integrates observations across locations through two components.
First, the Multi-View Bridging Module (MVBM) synthesizes interpolation videos between distant camera locations, increasing effective view overlap and strengthening cross-view geometric constraints.
Second, the Multi-Video Joint Optimization Module (MVJOM) jointly optimizes real panoramic observations and synthesized interpolation videos within a shared time-varying representation.
This design improves inter-location geometric consistency and stabilizes reconstruction across viewpoints and timestamps, yielding a spatially aligned 4D scene reconstruction from sparse, widely separated observations.

%
To enable systematic evaluation of sparse multi-location reconstruction, we further introduce Urban Sparse 4D (U-S4D), a controlled benchmark built on the CARLA driving simulator~\cite{Dosovitskiy17}.
U-S4D provides multi-location panoramic observations and standardized protocols for assessing spatial alignment in dynamic urban scenes.

The main contributions are summarized as follows:
\begin{itemize}
    \item[$\bullet$] We formulate SP4DR, a sparse multi-location 4D reconstruction problem that aims to recover a unified spatiotemporal representation from geographically separated videos.
    \item[$\bullet$] We propose \proposed, a unified framework for SP4DR comprising (i)~MVBM, which generates geometrically consistent interpolation videos between distant viewpoints, and (ii)~MVJOM, which jointly optimizes real and synthesized observations within a shared 4D representation.
    \item[$\bullet$] We introduce U-S4D, a controlled CARLA-based benchmark for SP4DR in dynamic urban scenes, with a unified evaluation protocol and metrics.
\end{itemize}

\vspace{-3mm}
\section{Related Work}
\vspace{-3mm}
\label{sec:related}

\para{Dynamic 4D Scene Reconstruction}
The aim of 4D scene reconstruction is to recover a temporally consistent representation of a dynamic environment from multi-view visual observations~\cite{cao2025reconstructing, kong20253d}. 
Early methods extended neural volumetric rendering to dynamic scenes by introducing time-dependent radiance fields~\cite{li2022streaming, nerfplayer23, attal2023hyperreel, fridovich2023k, li2022dynerf, Kumar_2025_CVPR}. 

Recent advances have shifted toward explicit primitive-based representations for efficient rendering, exemplified by 3D Gaussian Splatting (3DGS)~\cite{kerbl20233dgs}.
Building on 3DGS, 4D Gaussian Splatting (4DGS) represents scene dynamics using time-varying parameters and supports real-time rendering~\cite{wu20244dgs}.
Spacetime Gaussian Feature Splatting (SpacetimeGS) augments Gaussian primitives with spatiotemporal features to model appearance and motion variations under the same rasterization~\cite{li2024spacetime}.
FreeTime Gaussian Splatting (FreeTimeGS) further introduces a continuous temporal parameterization of dynamic Gaussians, enabling flexible temporal interpolation~\cite{wang2025freetimegs}.

Despite these advances, most dynamic reconstruction methods assume densely overlapping captures within a single area or object-centric sequences~\cite{wu20244dgs, li2024spacetime, wang2025freetimegs, chen2025dash, yuan2025, song2025coda, Zhang_2025_ICCV, park2025splinegs, yan2025instant, li20254d, gao20257dgs}.
While sparse-view reconstruction has been studied~\cite{Sparse4DGS25, yang2025storm, mihajlovic2024splatfields, wang2025monofusion, chen2024mvsplat}, existing methods mainly relax viewpoint density and do not address large, spatially separated camera layouts.
In contrast, \proposed~targets sparse multi-location 4D reconstruction by synthesizing camera-conditioned intermediate observations and jointly optimizing real and synthesized views in a shared 4D representation.

\vspace{-1mm}
\para{Urban Scene Modeling}
Through scalable training strategies and efficient rendering representations~\cite{blocknerfcvpr22, kerbl20233dgs, xiangli2022bungeenerf, Turki_2022_CVPR}, urban scene modeling has progressed. 
The resulting urban 3D assets support city-level downstream tasks such as navigation and geography-aware querying~\cite{Lee_2025_ICCV, yasuki2025geoprog3d, miyanishi_2023_NeurIPS, lee2025citynav}. 
Although static urban reconstructions have been extensively studied, the modeling of dynamic urban scenes with temporal consistency remains underexplored.

Autonomous-driving benchmarks such as Waymo, nuScenes, and KITTI provide rich dynamic street scenes~\cite{waymo, caesar2020nuscenes, kittiGeiger2012CVPR, Liao2022PAMI}.
They have enabled numerous methods for urban reconstruction under real-world conditions~\cite{yan2024street, chen2025omnire, wei2025emd, huang2026s3gaussian, song2025coda, Peng_2025_CVPR, chen2025snerf, Zhou_2024_CVPR}.
Despite this progress, existing benchmarks and street-view reconstruction pipelines do not systematically provide multi-view synchronized observations, or dense ground-truth images at arbitrary viewpoints, which are essential for training and evaluating free-viewpoint novel-view synthesis~\cite{waymo, caesar2020nuscenes, kittiGeiger2012CVPR, kirillov2019panoptic, xie2023snerf, yan2024street, sun2025splatflow, xu2025ad}.

Urban environments are often captured using panoramic videos, which provide full-view observations of complex scenes with multiple dynamic agents.
Accordingly, several panoramic-video datasets for urban environments, such as Leader360V and 360+x, have been proposed~\cite{zhang2025leaderv, chen2024x360}.
While these datasets provide panoramic captures, they are not designed for multi-location, time-synchronized 4D reconstruction with standardized free-viewpoint evaluation.
As a result, existing panoramic-video settings do not systematically address how to
(i) temporally align panoramic videos from multiple sparse locations,
(ii) integrate them into a unified dynamic 4D representation, or
(iii) train and evaluate novel-view synthesis under a systematically defined free-viewpoint evaluation protocol~\cite{zhang2025leaderv, chen2024x360}.
Nevertheless, panoramic videos remain a natural sensing modality for urban environments due to their full-view coverage.

In summary, existing benchmarks and methods largely assume single-location capture or densely overlapping views, leaving 4D reconstruction from spatially distributed urban locations underexplored.

\vspace{-3mm}
\section{Problem Formulation}
\vspace{-2mm}
We introduce the SP4DR problem, which reconstructs a unified spatiotemporal 4D representation of dynamic urban scenes from videos captured at spatially separated locations.
In this work, we focus on panoramic videos, which naturally arise in urban deployments such as surveillance systems and vehicle-mounted 360$^\circ$ cameras.
The representation must maintain geometric and temporal consistency while enabling novel-view rendering from arbitrary viewpoints and timestamps.
The left part of Fig.~\ref{fig:eye-catch} illustrates this sparse multi-location setting.

Formally, let \(\{V_i\}_{i=1}^{N_v}\) denote the set of panoramic videos captured at known camera locations \(p_i=(x_i,y_i,z_i)\) in a shared global coordinate system.
The objective is to estimate a time-varying scene representation $\mathcal{G}(t)$ that supports consistent rendering across viewpoints and time.

\vspace{-3mm}
\section{Methodology}
\vspace{-2mm}
\label{sec:method}
We propose \proposed, a framework that reconstructs urban scene dynamics as a continuous 4D representation from sparse multi-location observations.
The central idea is to complement missing spatiotemporal observations by generating camera-conditioned videos that bridge spatial gaps between panoramic viewpoints.
%
Our framework is built on existing 4D reconstruction methods~\cite{wu20244dgs, li2024spacetime, wang2025freetimegs}.
\proposed~bridges missing inter-location observations through generative completion and is agnostic to the underlying 4D representation and optimization strategy (see supplementary material).
In this work, we adopt the framework using SpacetimeGS~\cite{li2024spacetime} as the reconstruction backbone.

\begin{figure*}[t]
    \centering
    \vspace{-4mm}
    \includegraphics[width=0.8\linewidth]{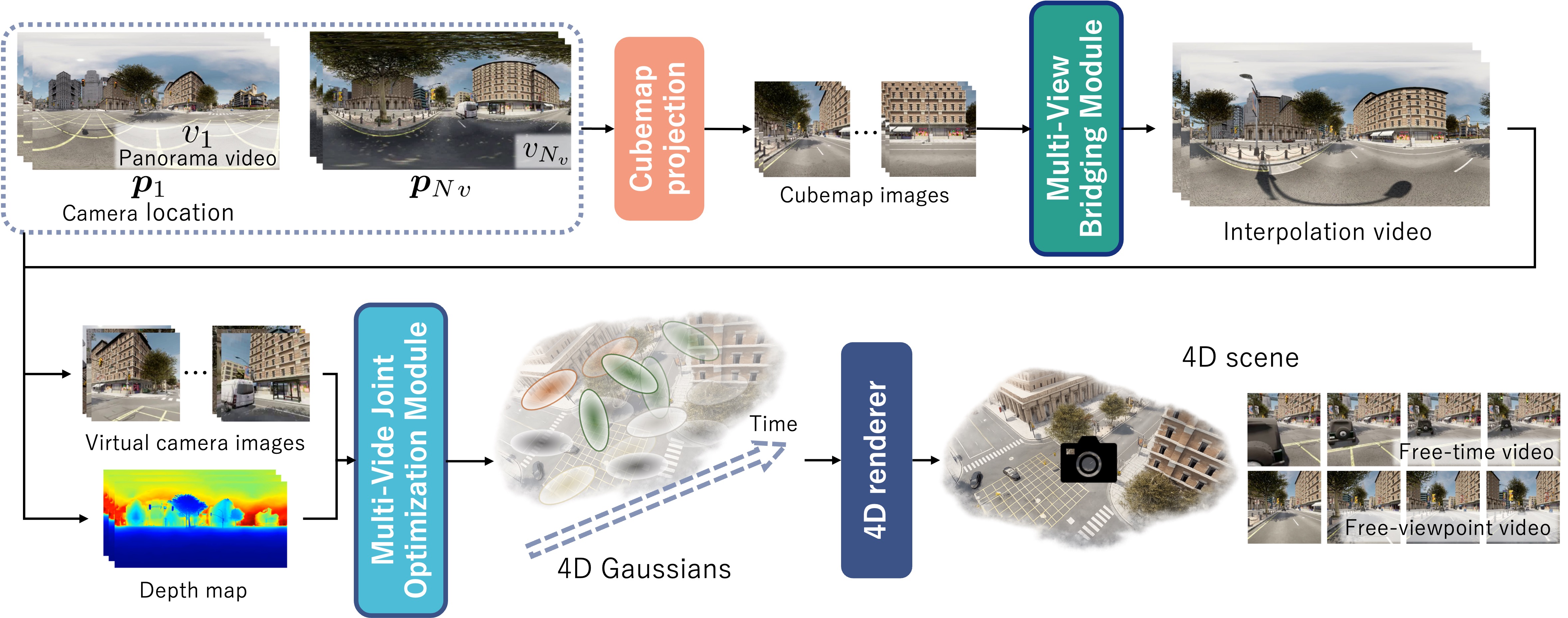}
    \vspace{-2mm}
    \caption{
    \textbf{Overall architecture of Stitch4D.}
    %
    MVBM synthesizes intermediate panoramic observations between spatially separated camera locations, while MVJOM jointly optimizes real and interpolated views to reconstruct a unified time-varying 4D representation.
    }
    \vspace{-5mm}
    \label{fig:framework}
\end{figure*}


\vspace{-2mm}
\subsection{Overview}
\vspace{-2mm}
The proposed method consists of two main modules: MVBM and MVJOM.
Fig.~\ref{fig:framework} shows the structure of~\proposed.
{\moduleone} converts a disconnected multi-location capture
into a weakly connected observation graph by inserting camera-conditioned
bridge observations between distant endpoints. 
{\moduletwo} then optimizes
real and bridge observations in a single coordinate frame, preventing
each endpoint from collapsing into an independent local reconstruction.

\vspace{-2mm}
\subsection{Data Representation}
\vspace{-2mm}
Formally, the input to \proposed~is defined as 
$\bm{x} = \{(\bm{v}_i, \bm{p}_i)\}_{i=1}^{N_v}$, 
where $\bm{v}_i \in \mathbb{R}^{T \times H \times W \times C}$ denotes the $i$-th panoramic video and $\bm{p}_i \in \mathbb{R}^3$ denotes its camera location in a shared world coordinate system. 
Here, $T$, $H$, $W$, and $C$ represent the number of frames, height, width, and channels, respectively.
For each $\bm{v}_i$, we estimate depth maps for all frames from RGB inputs using a monocular depth estimator.
The same estimated depths are used for all methods; simulator-provided ground-truth depth is not used.
Specifically, we define the set of cubemaps as~$\mathcal{C} = \{\bm{C}_i \mid i \in \mathcal{N}_v\}$, 
where $\bm{C}_i = \{\bm{c}_i^d \mid d \in \mathcal{D}\}$ consists of six directional perspective videos with 
$\mathcal{D} = \{\text{front}, \text{back}, \text{left}, \text{right}, \text{up}, \text{down}\}$.

To improve spatial coverage, we convert each panoramic video into a set of perspective views.
Specifically, we place virtual cameras on the unit sphere along 20 uniform directions and extract multiple views per direction with angular offsets. 
In addition, we introduce 20 wide field-of-view cameras to capture broader spatial context. 
Consequently, each $\bm{v}_i$ is converted into a set of perspective views 
$\mathcal{U}_i = \{\bm{u}_i^{(v)} \mid v \in \mathcal{V}\}$ with $|\mathcal{V}| = 120$. 
These views provide dense multi-directional observations that better capture spatial structure and temporal dynamics in urban scenes. 
Details are in the Supplementary Material.

\vspace{-2mm}
\subsection{Backbone}
\vspace{-2mm}
As the underlying 4D representation, we adopt SpacetimeGS, which models dynamic scenes using time-varying Gaussian primitives and supports efficient differentiable rendering. 
This representation serves as the optimization backbone of \proposed.
%
SpacetimeGS extends 3D Gaussian Splatting to the temporal domain, enabling continuous modeling of scene dynamics.
At time $t \in [0, T]$, the scene is represented as a set of time-varying Gaussian primitives as follows:
\begin{equation}
\setlength{\abovedisplayskip}{3pt}
\setlength{\belowdisplayskip}{3pt}
\mathcal{G}(t) = \{(\bm{\mu}_i(t), \bm{\Sigma}_i(t), \sigma_i(t), \bm{f}_i(t))\}_{i=1}^{N_{\mathrm{gp}}},
\end{equation}
where $N_{\mathrm{gp}}$ denotes the number of primitives.
For the $i$-th primitive at time $t$, $\bm{\mu}_i(t) \in \mathbb{R}^3$ and $\bm{\Sigma}_i(t) \in \mathbb{R}^{3\times3}$ represent its 3D mean and covariance, $\sigma_i(t)$ denotes the opacity coefficient, and $\bm{f}_i(t)$ is a learnable appearance feature.


Given a camera and time $t$, each Gaussian is projected onto the image plane through the camera model.
Let $\bar{\bm{\mu}}_i(t) \in \mathbb{R}^2$ and $\bar{\bm{\Sigma}}_i(t) \in \mathbb{R}^{2\times2}$ denote the projected mean and covariance obtained by applying the perspective projection and Jacobian-based covariance transformation to $\bm{\mu}_i(t)$ and $\bm{\Sigma}_i(t)$.
The contribution of the $i$-th Gaussian at pixel location $\bm{u} \in \mathbb{R}^2$ is defined as:
\begin{equation}
\setlength{\abovedisplayskip}{3pt}
\setlength{\belowdisplayskip}{3pt}
\resizebox{0.9\linewidth}{!}{$
\displaystyle
\alpha_i(\bm{u}, t)
=
\sigma_i(t)\,
\exp\!\left(
-\frac{1}{2}
(\bm{u}-\bar{\bm{\mu}}_i(t))^{\top}
\bar{\bm{\Sigma}}_i(t)^{-1}
(\bm{u}-\bar{\bm{\mu}}_i(t))
\right)
$}
\end{equation}
The pixel value $\mathbf{I}(\bm{u}, t)$ is obtained by $\alpha$-blending the depth-sorted Gaussians:
\begin{equation}
\setlength{\abovedisplayskip}{3pt}
\setlength{\belowdisplayskip}{3pt}
\mathbf{I}(\bm{u}, t)
=
\sum_{i \in \mathcal{N}(\bm{u}, t)}
\bm{r}_i(t)\,
\alpha_i(\bm{u}, t)
\prod_{j<i}
\left(1 - \alpha_j(\bm{u}, t)\right),
\end{equation}
where $\mathcal{N}(\bm{u}, t)$ denotes the set of Gaussians contributing to pixel $\bm{u}$ at time $t$, ordered by depth, and $\bm{r}_i(t)$ is the time-dependent color derived from the appearance feature $\bm{f}_i(t)$.


\vspace{-2mm}
\subsection{Multi-View Bridging Module}
\vspace{-1mm}

MVBM synthesizes panoramic videos at intermediate viewpoints to improve geometric and temporal consistency in 4D reconstruction from multi-location observations.
As shown in Fig.~\ref{fig:framework}, panoramic videos captured at spatially separated locations often exhibit limited viewpoint overlap.
This leads to unstable geometric correspondences and discontinuous visibility, especially under sparse-view configurations with large inter-location gaps.
By synthesizing intermediate observations, MVBM increases view overlap across locations and strengthens geometric constraints, promoting globally consistent 4D reconstruction.

\noindent\textbf{Panoramic Video Interpolation.}
%
%
The input to MVBM is the set of cubemap video groups $\mathcal{C}$.
From $\mathcal{C}$, we select two cubemap video groups $\bm{C}_i$ and $\bm{C}_j$ captured at camera locations $\bm{p}_i$ and $\bm{p}_j$, where $i \neq j$.
We then generate a set of interpolated panoramic videos,
$\hat{\mathcal{V}} = \{ \bm{v}_k \mid k \in \mathcal{K} \}$,
where $\mathcal{K}$ denotes intermediate camera locations uniformly sampled along the line segment connecting $\bm{p}_i$ and $\bm{p}_j$.
For each interpolated location $\bm{p}_k$, panoramic frames are synthesized independently at each timestamp and temporally assembled to form an interpolation video.

\begin{figure*}[t]
    \centering
    \vspace{-4mm}
    \includegraphics[width=0.8\linewidth]{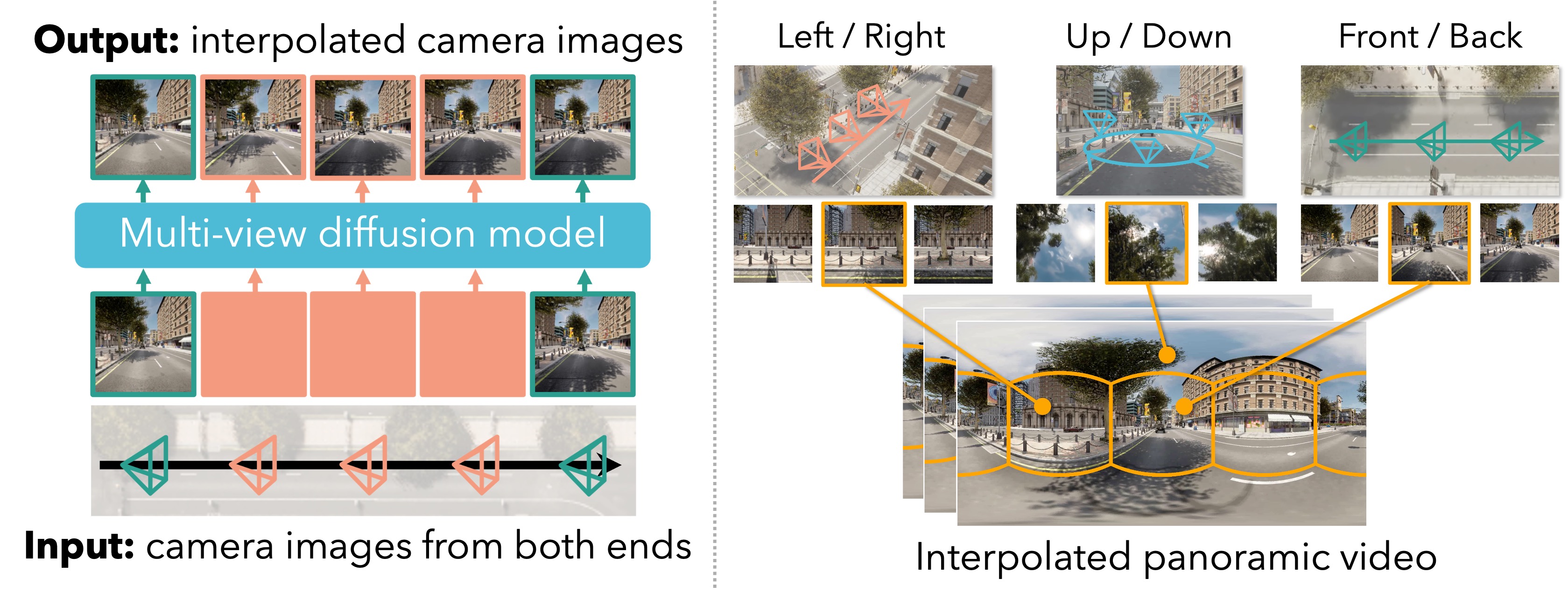}
    \vspace{-6mm}
    \caption{
        \textbf{Overview of MVBM.}
        MVBM synthesizes intermediate panoramic videos between spatially separated cameras.
        It first generates intermediate views with a multi-view diffusion model (left) and then assembles them into panoramic videos (right).
    }
    \vspace{-6mm}
    \label{fig:mvbm}
\end{figure*}

\noindent\textbf{Interpolated Panoramic Image Generation.}
Fig.~\ref{fig:mvbm} shows the pipeline for generating interpolated panoramic images.
Given two cubemap video groups $\bm{C}_i$ and $\bm{C}_j$, we synthesize intermediate views using 
a camera-conditioned multi-view diffusion model~\cite{zhou2025stable}.
The model generates intermediate frames conditioned on camera poses sampled along the trajectory between the two input viewpoints.
Interpolation is performed between the geometrically corresponding cubemap faces of $\bm{C}_i$ and $\bm{C}_j$ for each of the four horizontal directions (front, back, left, and right).
%
For each interpolated location $\bm{p}_k$ and timestamp $t$, a panoramic frame is synthesized from the corresponding input frames at time $t$.
To recover the full field of view, additional virtual views for the up and down directions are rendered by rotating the camera along the pitch axis while keeping the camera center fixed.
The six directional images are assembled into a cubemap and reprojected into equirectangular format to produce a panoramic frame at $\bm{p}_k$.
Repeating this process for all $k \in \mathcal{K}$ yields a spatially interpolated video sequence that bridges the gap between $\bm{p}_i$ and $\bm{p}_j$.

\vspace{-2mm}
\subsection{Multi-Video Joint Optimization Module}
\vspace{-1mm}
MVJOM integrates panoramic videos captured from multiple camera locations and jointly optimizes a single time-varying scene representation $\mathcal{G}(t)$ in a unified coordinate frame.
Conventional 4D reconstruction methods optimize each location independently without explicitly enforcing geometric consistency across locations~\cite{li2024spacetime, wu20244dgs, wang2025freetimegs}.
Under sparse-view configurations with limited viewpoint overlap, this decoupled optimization results in insufficient geometric constraints.
As a result, geometry estimation and dynamic modeling become unstable, particularly in transition regions between camera locations.

MVJOM takes as input a set of perspective images $\mathcal{U}_i$, corresponding depth maps, and camera locations, and minimizes the reconstruction error between the observed images and renderings of $\mathcal{G}(t)$ from each viewpoint.
In addition to photometric consistency, MVJOM introduces regularization terms (see Section~\ref{sec:lossfunc}) to enforce inter-location spatial consistency and stabilize reconstruction across timestamps.
These constraints enable the estimation of a spatially aligned and stable 4D scene representation across viewpoints and timestamps.

\vspace{-2mm}
\subsection{Seam-Aware Inter-Location Regularization}
\vspace{-1mm}
\label{sec:lossfunc}
Limited viewpoint overlap across disjoint camera locations makes dense cross-view supervision unreliable.
Relying only on standard photometric reconstruction can leave neighboring locations weakly coupled, especially around transition regions.
We use a seam-aware regularization term around camera-location boundaries:
\begin{equation}
\setlength{\abovedisplayskip}{3pt}
\setlength{\belowdisplayskip}{3pt}
\mathcal{L}_{\mathrm{SAIL}}
=
\beta(\delta)\,\mathcal{L}_{\mathrm{recon}}
+
\lambda_{\mathrm{inter}}\,\mathcal{L}_{\mathrm{inter}} .
\end{equation}
Here, $\mathcal{L}_{\mathrm{recon}}$ denotes the photometric reconstruction loss, and $\beta(\delta)$ increases its weight near camera-location boundaries according to the distance $\delta$ to the nearest boundary.
The scalar $\lambda_{\mathrm{inter}}$ controls the contribution of the inter-location term.
The inter-location term provides a local consistency cue between neighboring locations:
\begin{equation}
\setlength{\abovedisplayskip}{3pt}
\setlength{\belowdisplayskip}{2pt}
\mathcal{L}_{\mathrm{inter}}
=
\gamma(\delta)\,
\left\|
\nabla \hat{\mathbf{I}}_1
-
\nabla \hat{\mathbf{I}}_2
\right\|_1 ,
\end{equation}
where $\hat{\mathbf{I}}_1$ and $\hat{\mathbf{I}}_2$ denote renderings of the same boundary region from adjacent locations, and $\nabla(\cdot)$ computes spatial image gradients via finite differences.
Here, $\gamma(\delta)$ is a boundary-aware weighting function that activates the inter-location term only near connection boundaries.
The inter-location term is applied only near connection boundaries between adjacent camera locations, where $\gamma(\delta)$ promotes local seam consistency without dense pixel-wise view alignment.
Detailed definitions of $\mathcal{L}_{\mathrm{recon}}$, $\beta(\delta)$, and $\gamma(\delta)$ are provided in the Appendix Section C.
\vspace{-3mm}
\section{Experimental Results}
\vspace{-2mm}
\label{sec:exp}
\subsection{Experimental Setup}
\vspace{-1mm}

\noindent\textbf{Controlled Benchmark.}
We constructed U-S4D, a controlled benchmark for sparse multi-location 4D reconstruction from synchronized panoramic videos in urban scenes.
Existing urban panoramic video datasets~\cite{zhang2025leaderv, chen2024x360} provide valuable 360$^\circ$ recordings, but they are not designed for sparse multi-location 4D reconstruction, which requires synchronized observations from spatially separated viewpoints and ground-truth intermediate views for controlled free-viewpoint evaluation.
U-S4D addresses this gap by providing synchronized multi-location panoramic videos with calibrated camera poses.
Rather than aiming for large-scale data coverage, U-S4D is designed to isolate sparse spatial coverage, weak inter-location overlap, and free-viewpoint reconstruction in unobserved intermediate regions.

\begin{figure}[!t]
    \centering
    \includegraphics[width=1\linewidth]{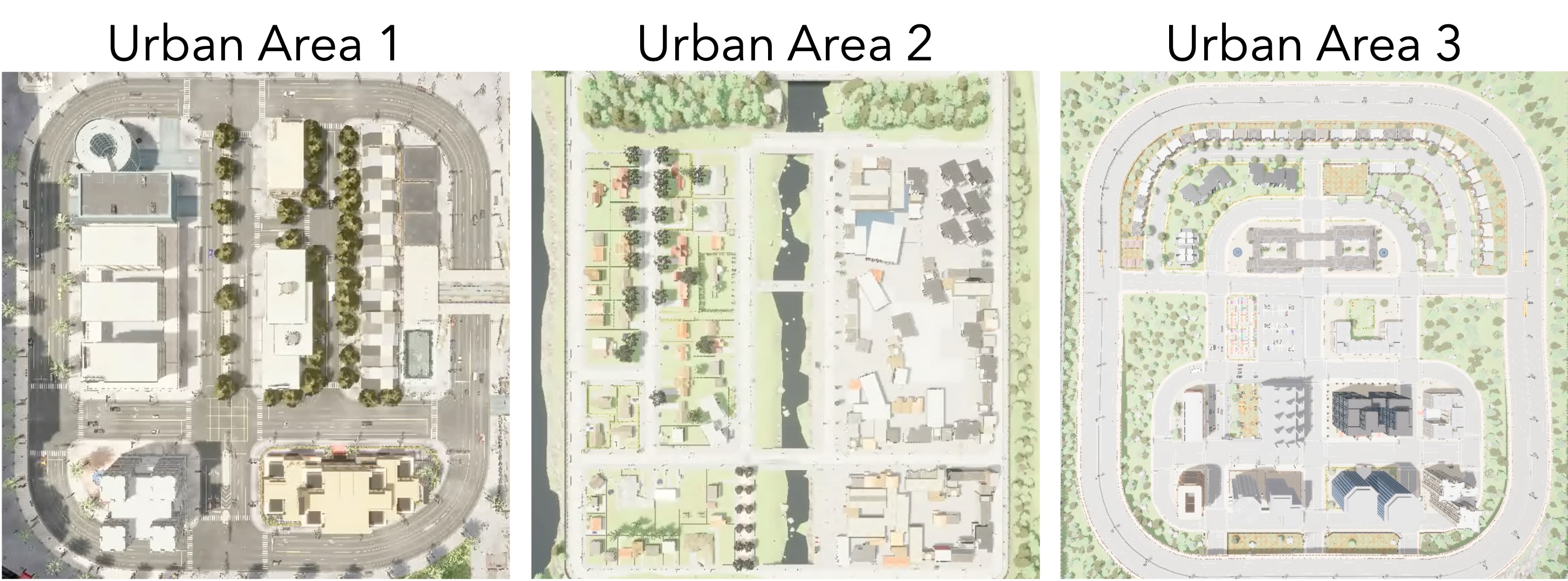}
    \vspace{-7mm}
    \caption{
     \textbf{Overview of U-S4D.}
    U-S4D consists of three CARLA urban areas used to construct sparse multi-location panoramic video scenes.
    }
    \vspace{-7mm}
    \label{fig:u-s4d}
\end{figure}

\vspace{-0.5mm}
\noindent\textbf{Dataset Construction.}
We built U-S4D using the autonomous driving simulator CARLA~\cite{Dosovitskiy17}, which enables precise control over dynamic urban environments and provides realistic traffic participants, including vehicles and pedestrians.
Fig.~\ref{fig:u-s4d} shows an overview of the scenes included in U-S4D.
As collecting synchronized multi-location observations with accurate camera poses in real-world urban environments is challenging, we construct U-S4D in simulation to enable controlled evaluation.
For each urban scene, we define spatially separated camera locations and capture temporally synchronized panoramic videos with ground-truth camera poses provided by the simulator.
We additionally place virtual viewpoints along continuous trajectories between camera locations to obtain ground-truth renderings for free-viewpoint novel-view synthesis.
This controlled setup enables systematic evaluation of reconstruction methods under the SP4DR setting.

\vspace{-0.5mm}
\noindent\textbf{Dataset Statistics.}
U-S4D consists of three urban areas, each containing two scenes, resulting in six scenes in total.
%
Each scene contains two synchronized panoramic videos captured from spatially separated camera locations within the corresponding urban area, together with their camera poses. 
The scenes span three urban environments: (i) Urban Area 1, a dense urban intersection with frequent vehicle and pedestrian traffic ($\approx 4.7 \times 10^{4}\,\mathrm{m}^2$); (ii) Urban Area 2, a residential district with high-rise buildings and narrow streets ($\approx 1.3 \times 10^{5}\,\mathrm{m}^2$); and (iii) Urban Area 3, a peri-urban arterial road with long-range visibility ($\approx 2.0 \times 10^{5}\,\mathrm{m}^2$).
Each panoramic video is 1~s long, captured at 10~fps, with a resolution of $512 \times 1024$.


\vspace{-0.5mm}
\noindent{\textbf{Evaluation Protocol.}}
%
%
Following standard 4D reconstruction evaluation protocols~\cite{li2024spacetime, wu20244dgs, wang2025freetimegs},
we consider two training settings: \emph{full reconstruction} and \emph{temporal split}.
Each setting is evaluated under two conditions, \emph{trajectory interpolation} and \emph{seen-viewpoints}, resulting in four evaluation configurations.
Ground-truth renderings from virtual trajectory viewpoints are used only for evaluation and never for optimization. 
The evaluation trajectory camera poses are disjoint from both the original input poses and the bridge camera poses used by MVBM, so that trajectory interpolation evaluates novel camera poses that are never used during optimization.



\noindent{\textbf{-- Full reconstruction setting}}:
This setting measures reconstruction fidelity when complete spatiotemporal observations are available.
All original input viewpoints and timestamps were used for training.

\noindent{\textbf{-- Temporal split setting}}:
Every fourth frame was held out for testing, and the remaining frames (1920 samples per scene) were used for training.
The held-out frames are used to evaluate novel-view synthesis at unseen timestamps and measure temporal generalization.

\noindent{\textbf{-- Trajectory interpolation condition}}:
This condition evaluates spatial generalization to intermediate regions that are not directly observed during training.
Virtual viewpoints were placed at 1\,m intervals along trajectories connecting camera locations.
We used 110 test samples for evaluation.

\noindent{\textbf{-- Seen-viewpoints condition}}:
This condition isolates temporal generalization while keeping spatial viewpoints fixed.
For the full reconstruction setting, performance was evaluated at all original camera viewpoints.
Under the temporal split setting, the 480 held-out timestamp samples constituted the test set.
For each scene, we trained each method on the corresponding training split and report the performance on the designated test split.

\vspace{-0.5mm}
\noindent\textbf{Baselines.}
We compared Stitch4D on SP4DR with two groups of representative reconstruction baselines:
urban-scene reconstruction methods, including Periodic Vibration Gaussian (PVG)~\cite{chen2026periodic} and Street Gaussians~\cite{yan2024street},
and general 4D reconstruction methods, including 4DGS~\cite{wu20244dgs}, SpacetimeGS~\cite{li2024spacetime}, and FreeTimeGS~\cite{wang2025freetimegs}.


\vspace{-0.5mm}
\noindent\textbf{Implementation Details.}
All experiments were conducted on a single NVIDIA H200 SXM GPU (141\,GB VRAM).
All baselines used the same panoramic-to-perspective conversion, camera poses, training frames, estimated depth maps, and evaluation protocol as \proposed.
The training time for the proposed model per scene was approximately 1 hour.
At a resolution of 512×512, the rendering time was 0.73~ms per image.
Further implementation details are provided in the Supplementary Materials.

\vspace{-0.5mm}
\noindent\textbf{Evaluation Metrics.}
We evaluate methods using standard image-quality metrics: Peak Signal-to-Noise Ratio (PSNR), Structural Similarity Index Measure (SSIM), and Learned Perceptual Image Patch Similarity (LPIPS), with PSNR used as the primary metric~\cite{wu20244dgs, li2024spacetime, wang2025freetimegs}.
\begin{table*}[t!]
\centering
\vspace{-4mm}
\caption{
Quantitative comparison with baseline methods.
The best score for each metric is in \textbf{bold}.
}
\vspace{-3mm}
\resizebox{\textwidth}{!}{%
\begin{tabular}{lcccccccccccc}
\toprule
\multirow{3}{*}{Method}
& \multicolumn{6}{c}{Full reconstruction}
& \multicolumn{6}{c}{Temporal split} \\
\cmidrule(lr){2-7}
\cmidrule(lr){8-13}
& \multicolumn{3}{c}{Trajectory interpolation}
& \multicolumn{3}{c}{Seen-viewpoints}
& \multicolumn{3}{c}{Trajectory interpolation}
& \multicolumn{3}{c}{Seen-viewpoints} \\
\cmidrule(lr){2-4}
\cmidrule(lr){5-7}
\cmidrule(lr){8-10}
\cmidrule(lr){11-13}
& \small PSNR $\uparrow$
& \small SSIM $\uparrow$
& \small LPIPS $\downarrow$
& \small PSNR $\uparrow$
& \small SSIM $\uparrow$
& \small LPIPS $\downarrow$
& \small PSNR $\uparrow$
& \small SSIM $\uparrow$
& \small LPIPS $\downarrow$
& \small PSNR $\uparrow$
& \small SSIM $\uparrow$
& \small LPIPS $\downarrow$ \\
\midrule

\rowcolor{gray!12}
\multicolumn{13}{l}{\textbf{Urban scene reconstruction methods}} \\

\makecell[tl]{PVG \cite{chen2026periodic}}
& 12.84 & 0.58 & 0.76
& 13.76 & 0.69 & 0.57
& 12.67 & 0.57 & 0.77
& 13.81 & 0.69 & 0.58 \\

\makecell[tl]{Street Gaussians \cite{yan2024street}}
& 11.85 & 0.56 & 0.75
& 16.18 & 0.72 & 0.54
& 11.49 & 0.53 & 0.75
& 17.38 & 0.73 & 0.51 \\

\rowcolor{gray!12}
\multicolumn{13}{l}{\textbf{General 4D reconstruction methods}} \\

\makecell[tl]{4DGS \cite{wu20244dgs}}
& 11.51 & 0.28 & 0.84
& 15.79 & 0.58 & 0.84
& 10.54 & 0.25 & 0.80
& 13.78 & 0.52 & 0.64 \\

\makecell[tl]{SpacetimeGS \cite{li2024spacetime}}
& 12.91 & 0.55 & 0.67
& 17.75 & 0.79 & 0.33
& 12.47 & 0.54 & 0.69
& 17.49 & 0.78 & 0.33 \\

\makecell[tl]{FreeTimeGS \cite{wang2025freetimegs}}
& 11.75 & 0.52 & 0.75
& 16.70 & 0.71 & 0.41
& 11.84 & 0.53 & 0.74
& 16.53 & 0.71 & 0.41 \\

\midrule

\makecell[tl]{\proposed~(ours)}
& \textbf{15.31} & \textbf{0.60} & \textbf{0.51}
& \textbf{26.34} & \textbf{0.92} & \textbf{0.13}
& \textbf{14.88} & \textbf{0.59} & \textbf{0.52}
& \textbf{24.63} & \textbf{0.90} & \textbf{0.15} \\

\bottomrule
\end{tabular}%
    }
\vspace{-5mm}
\label{tab:quantitative}
\end{table*}

\vspace{-2mm}
\subsection{Quantitative Results}
\vspace{-2mm}

Table~\ref{tab:quantitative} reports the quantitative results on U-S4D in the full reconstruction and temporal split settings.
These results are reported as the average over all scenes in U-S4D.
For each metric, the best result is highlighted in \textbf{bold}.
\proposed~outperformed the baselines in terms of PSNR, SSIM, and LPIPS across all evaluation configurations.


In the full reconstruction setting, {\proposed} achieved a PSNR of 15.31\,dB under the trajectory interpolation condition and 26.34\,dB under the seen-viewpoints condition, outperforming the best baseline results of 12.91\,dB and 17.75\,dB, respectively.
In the temporal split setting, {\proposed} achieved 14.88\,dB (trajectory interpolation) and 24.63\,dB (seen-viewpoints), whereas the best baseline obtained 12.67\,dB and 17.49\,dB, respectively.
We observe consistent improvements across all configurations.

These results demonstrate the benefit of integrating multi-location observations into a unified time-varying representation.
By improving geometric consistency across spatially separated locations in a shared coordinate system,
{\proposed} generalizes to intermediate viewpoints and unseen timestamps, which are challenging under sparse wide-baseline camera layouts.



\vspace{-2mm}
\subsection{Qualitative Results}
\vspace{-2mm}
\begin{figure}[t]
    \centering
    \vspace{-2mm}
    \includegraphics[width=1\linewidth]{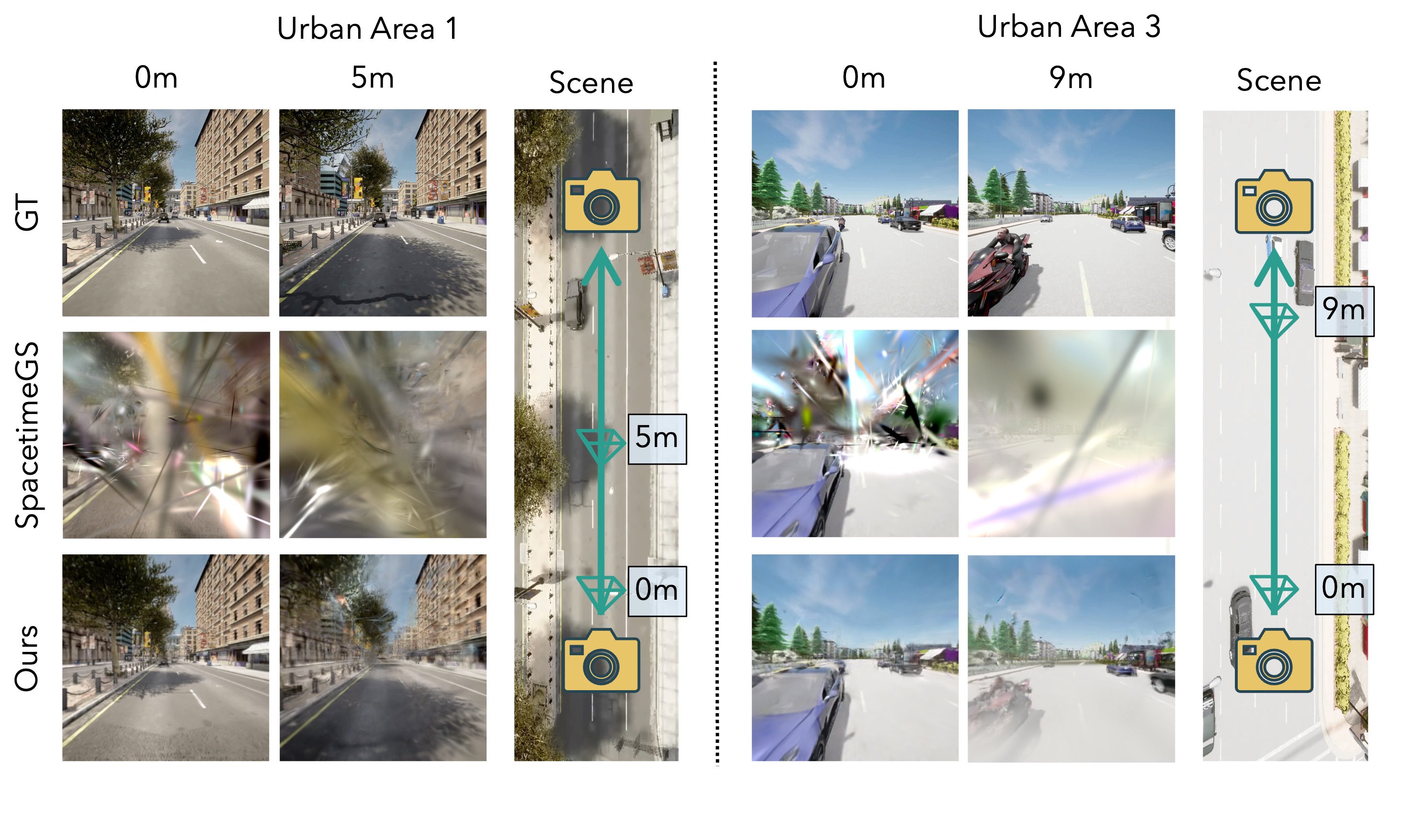}
    \vspace{-11mm}
    \caption{
    Qualitative results in the full reconstruction setting (trajectory interpolation condition). 
    Yellow cameras indicate input viewpoints, while green icons denote evaluation viewpoints. 
    }
    \vspace{-7mm}
    \label{fig:qual_recon_backforth}
\end{figure}
\begin{figure}[t]
    \centering
    \vspace{-2mm}
    \includegraphics[width=1\linewidth]{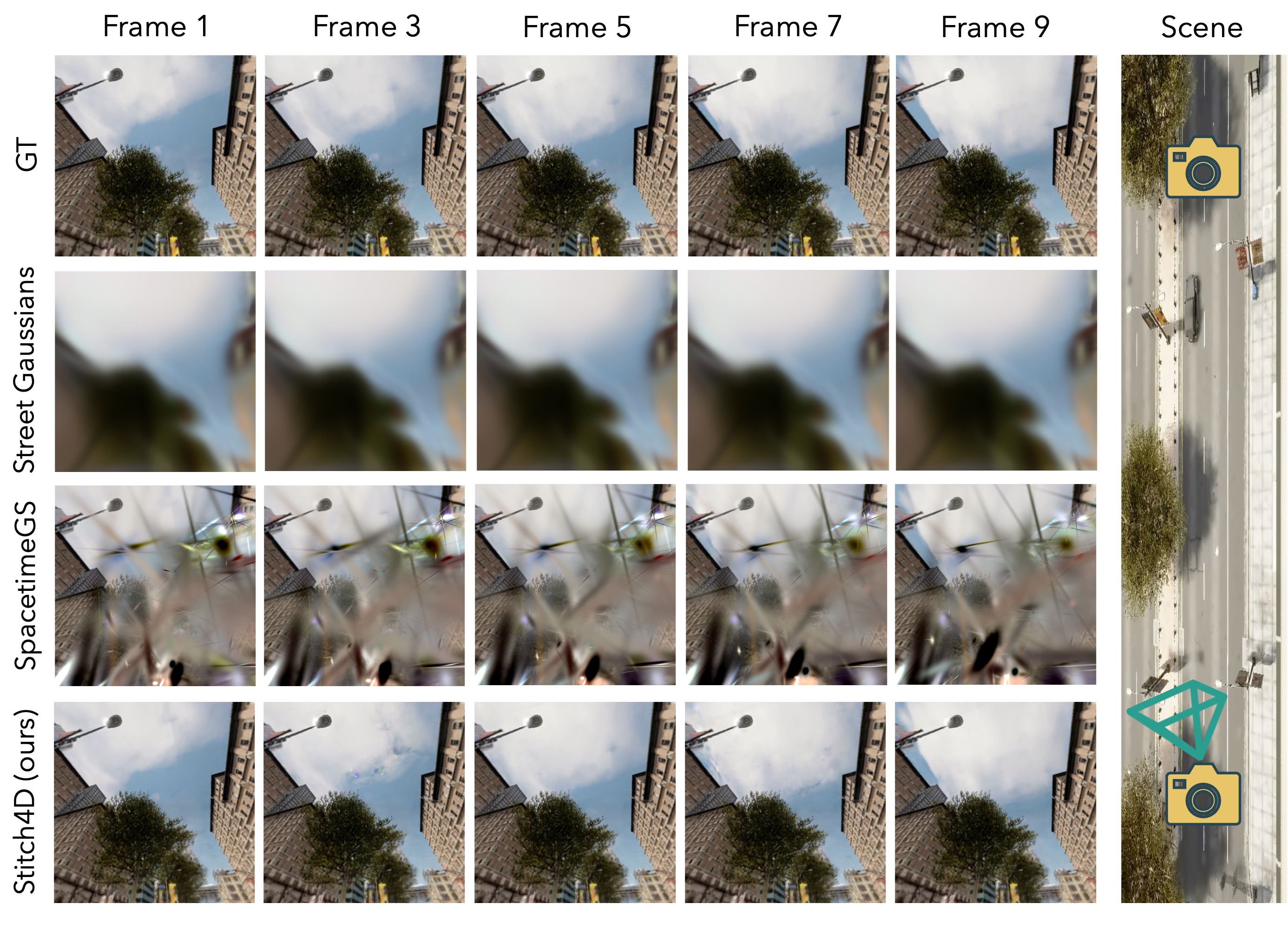}
    \vspace{-8mm}
    \caption{
    Qualitative results in the temporal split setting (seen-viewpoints condition) for Urban Area 1. 
    Each column denotes the frame index (test frames: 3, 7). 
    }
    \label{fig:qual_temporal_virtual_1}
    \vspace{-6mm}
\end{figure}


Figs.~\ref{fig:qual_recon_backforth} and \ref{fig:qual_temporal_virtual_1} show the qualitative results of Stitch4D
and the baseline methods.
Fig.~\ref{fig:qual_recon_backforth} shows the trajectory interpolation condition in the full reconstruction setting, whereas Fig.~\ref{fig:qual_temporal_virtual_1} shows the seen-viewpoints condition in the temporal split setting.
Further qualitative analyses are provided in the supplementary material.

In Fig.~\ref{fig:qual_recon_backforth}, each column renders the scene from the virtual viewpoint indicated at the top.
Because SpacetimeGS optimizes scene representations independently for each camera location, cross-location alignment is only weakly constrained.
As the viewpoint moved to intermediate positions, this resulted in geometric distortions, severe blurring, and occasionally divergent artifacts.
In contrast, our method jointly optimized observations from multiple locations together with interpolated viewpoints in a shared coordinate system.
This design maintains geometric consistency, producing stable renderings even at intermediate viewpoints.

In Fig.~\ref{fig:qual_temporal_virtual_1}, each row corresponds to a fixed virtual camera, and each column corresponds to the frame index shown at the top (test frames: 3, 7).
In Urban Area~1 (Fig.~\ref{fig:qual_temporal_virtual_1}), SpacetimeGS results exhibit structural blur and unstable artifacts on the held-out frames, resulting in ambiguous boundaries.
\proposed~preserves sharper boundaries (e.g., buildings versus sky) and produces more stable renderings across held-out frames.

\begin{figure}[!th]
    \centering
    \vspace{-1mm}
    \includegraphics[width=0.95\linewidth]{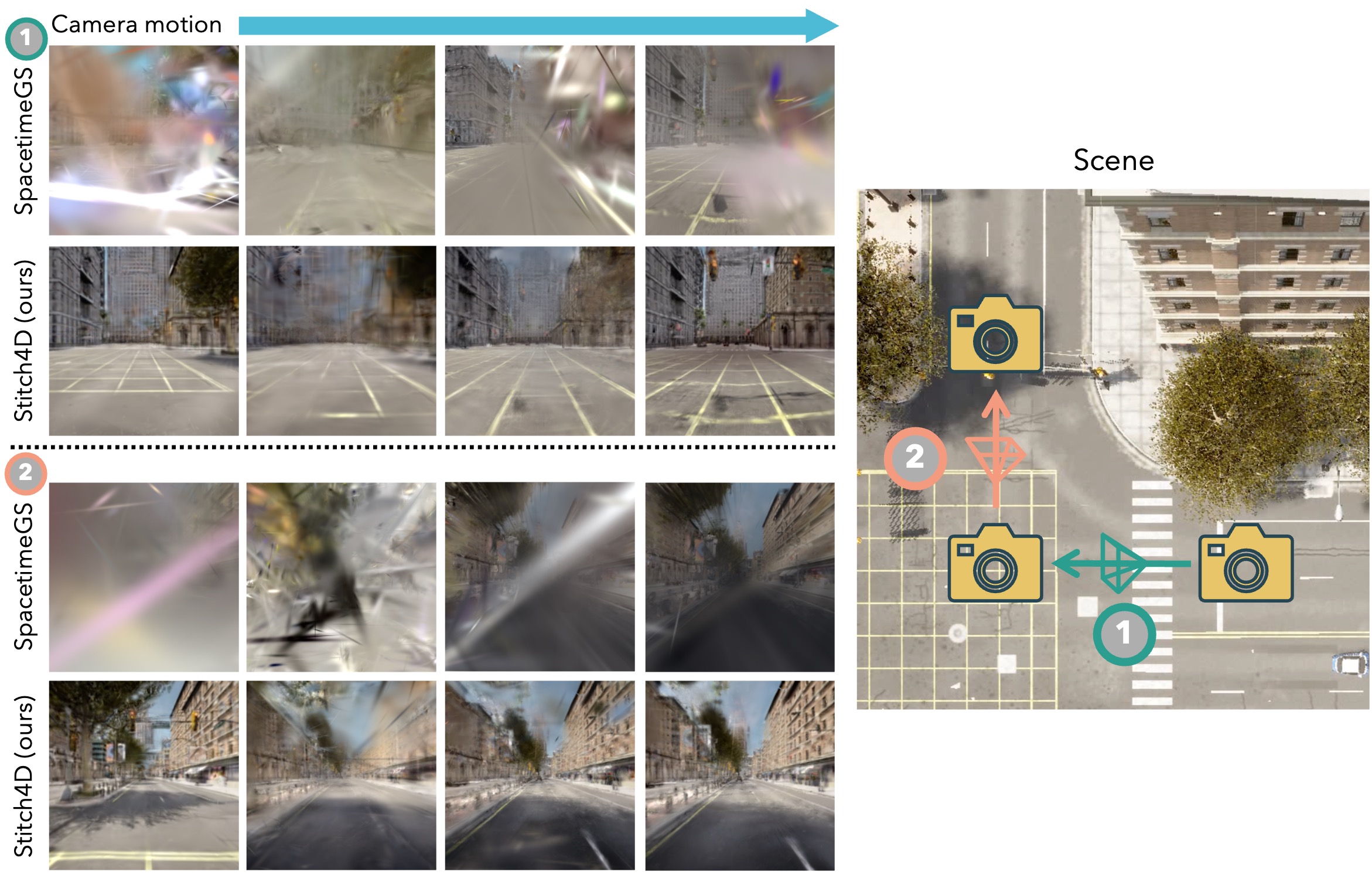}
    \vspace{-2mm}
    \caption{
    Qualitative comparison on Urban Area 2 under the full reconstruction setting with freely moving camera trajectories, reconstructed from three input videos.
    }
    \vspace{-2mm}
    \label{fig:3input_lbrf}
\end{figure}

We further evaluate the method with an increased number of input videos.
Fig.~\ref{fig:3input_lbrf} presents qualitative results obtained using three input videos, suggesting that the proposed framework can be extended beyond the two-input setting.
Additional qualitative results are provided in the supplementary material.

Overall, these examples show that \proposed~produces stable renderings in sparse urban observations.


\vspace{-2mm}
\subsection{Ablation Study}
\vspace{-1mm}
\label{sec:ablation}
Table~\ref{tab:ablation_1} shows the quantitative results of ablation studies. The best score for each metric is shown in \textbf{bold}.
Further ablations are provided in the supplementary material.

\begin{table}[t!]
\centering
\caption{Quantitative results of the ablation studies under the full reconstruction setting.}
\vspace{-4mm}
\resizebox{\columnwidth}{!}
{
\begin{tabular}{lcccccc}
\toprule
\multirow{2}{*}{Model}
& \multicolumn{3}{c}{Trajectory interpolation}
& \multicolumn{3}{c}{Seen-viewpoints}\\
\cmidrule(lr){2-4}
\cmidrule(lr){5-7}
& \small PSNR  $\uparrow$
& \small SSIM $\uparrow$
& \small LPIPS $\downarrow$
& \small PSNR $\uparrow$
& \small SSIM $\uparrow$
& \small LPIPS $\downarrow$
\\
\midrule 

\makecell[tl]{w/o MVBM}
& \makecell[tc]{14.51}
& \makecell[tc]{0.57}
& \makecell[tc]{0.61}
& \makecell[tc]{24.68}
& \makecell[tc]{0.88}
& \makecell[tc]{0.18}
\\

\makecell[tl]{w/o MVJOM, MVBM}
& \makecell[tc]{12.91} & \makecell[tc]{0.55} & \makecell[tc]{0.67}
& \makecell[tc]{17.75} & \makecell[tc]{0.79} & \makecell[tc]{0.33}
\\

\makecell[tl]{Stitch4D (ours)}
& \makecell[tc]{\textbf{15.31}}
& \makecell[tc]{\textbf{0.60}}
& \makecell[tc]{\textbf{0.51}}
& \makecell[tc]{\textbf{26.34}}
& \makecell[tc]{\textbf{0.92}}
& \makecell[tc]{\textbf{0.13}}
\\

\bottomrule
\end{tabular}
}
\vspace{-3mm}
\label{tab:ablation_1}
\end{table}

\begin{table}[t]
\centering
\caption{
Quantitative ablation results on robustness to camera-position noise proportional to the inter-location distance under the full reconstruction setting. }
\vspace{-2mm}
\setlength{\tabcolsep}{4.0pt}
\resizebox{0.99\columnwidth}{!}{%
\begin{tabular}{lcccccc}
\toprule
\multirow{2}{*}{Noise}
& \multicolumn{3}{c}{Trajectory interpolation}
& \multicolumn{3}{c}{Seen-viewpoints} \\
\cmidrule(lr){2-4}
\cmidrule(lr){5-7}
& PSNR $\uparrow$ & SSIM $\uparrow$ & LPIPS $\downarrow$
& PSNR $\uparrow$ & SSIM $\uparrow$ & LPIPS $\downarrow$ \\
\midrule
0\%  
& \textbf{15.31} & \textbf{0.60} & \textbf{0.51}
& \textbf{26.34} & \textbf{0.92} & \textbf{0.13} \\


5\%  
& 14.51 & 0.57 & 0.54
& \textbf{26.34} & \textbf{0.92} & \textbf{0.13} \\

10\% 
& 14.39 & 0.57 & 0.53
& 26.08 & \textbf{0.92} & 0.14 \\
\bottomrule
\end{tabular}
}
\vspace{-5mm}
\label{tab:noise}
\end{table}


\begin{table*}[t]
\centering
\caption{
Quantitative results on complementary real-world 360$^\circ$ video datasets under the full reconstruction setting.
}
\vspace{-3mm}
\resizebox{\textwidth}{!}{%
\begin{tabular}{@{}lcccccccccccc@{}}
\toprule
\multirow{3}{*}{Method}
& \multicolumn{6}{c}{Dur360BEV~\cite{dur360bev}}
& \multicolumn{6}{c}{FTV360~\cite{maugey2019ftv360}} \\
\cmidrule(lr){2-7}
\cmidrule(lr){8-13}
& \multicolumn{3}{c}{Trajectory interpolation}
& \multicolumn{3}{c}{Seen-viewpoints}
& \multicolumn{3}{c}{Trajectory interpolation}
& \multicolumn{3}{c}{Seen-viewpoints} \\
\cmidrule(lr){2-4}
\cmidrule(lr){5-7}
\cmidrule(lr){8-10}
\cmidrule(lr){11-13}
& \small PSNR $\uparrow$
& \small SSIM $\uparrow$
& \small LPIPS $\downarrow$
& \small PSNR $\uparrow$
& \small SSIM $\uparrow$
& \small LPIPS $\downarrow$
& \small PSNR $\uparrow$
& \small SSIM $\uparrow$
& \small LPIPS $\downarrow$
& \small PSNR $\uparrow$
& \small SSIM $\uparrow$
& \small LPIPS $\downarrow$ \\
\midrule

\makecell[tl]{SpacetimeGS \cite{li2024spacetime}}
& 10.44 & 0.48 & 0.76
& 10.26 & 0.51 & 0.60
& 13.88 & 0.56 & 0.75
& 11.31 & 0.58 & 0.72 \\

\makecell[tl]{FreeTimeGS \cite{wang2025freetimegs}}
& 9.45 & 0.44 & 0.81
& 10.41 & 0.49 & 0.59
& 13.90 & 0.55 & 0.71
& 10.52 & 0.55 & 0.73 \\
\midrule

\makecell[tl]{\proposed~(ours)}
& \textbf{16.03} & \textbf{0.57} & \textbf{0.47}
& \textbf{22.23} & \textbf{0.90} & \textbf{0.15}
& \textbf{19.32} & \textbf{0.66} & \textbf{0.48}
& \textbf{22.66} & \textbf{0.90} & \textbf{0.20} \\

\bottomrule
\end{tabular}%
}
\vspace{-5mm}
\label{tab:real-world}
\end{table*}

\noindent\textbf{Effect of MVBM.}
%
To assess the contribution of MVBM, we remove it from the full model. 
The resulting variant achieves 14.51\,dB and 24.68\,dB PSNR under the trajectory interpolation and seen-viewpoints conditions, respectively. 
This corresponds to drops of 0.80\,dB and 1.66\,dB compared with the full model. 
These results demonstrate that MVBM bridges camera location gaps, improving geometric consistency and reconstruction stability across viewpoints.

\noindent\textbf{Effect of MVJOM.}
%
To isolate the contribution of MVJOM, we compare the model without both MVJOM and MVBM against the model without MVBM.
Removing MVJOM reduces PSNR by 1.60\,dB under the trajectory interpolation condition and by 6.93\,dB under the seen-viewpoints condition in the full reconstruction setting. 
The larger drop in the seen-viewpoints condition indicates that joint optimization across locations is critical for aligning multi-location observations within a shared coordinate system. 
These results indicate that MVJOM improves 
geometric alignment and reconstruction stability across viewpoints.

\noindent\textbf{Robustness to inter-location distance noise.}

Table~\ref{tab:noise} shows the robustness results under camera-position noise. To evaluate robustness to calibration errors in sparse multi-location settings, we perturb the inter-location distance with zero-mean Gaussian noise whose standard deviation is set to 5\% and 10\% of the distance between camera locations.
Even under 10\% noise, which corresponds to a 1 m perturbation for a 10~m inter-location distance, the proposed method achieves 14.39 dB and 26.08 dB PSNR under the trajectory interpolation and seen-viewpoints conditions, respectively.
These correspond to drops of 0.92 dB and 0.26 dB compared with the full model without noise.
SSIM and LPIPS also remain stable, indicating that the proposed method is robust to inter-location distance perturbations.

\vspace{-2mm}
\subsection{Real-World Evaluation}
\label{sec:real-world}

\vspace{-2mm}
\noindent\textbf{Experimental setup.}
To examine whether Stitch4D transfers beyond the controlled CARLA benchmark, we further evaluate it on two complementary real-world 360$^\circ$ video datasets: Dur360BEV~\cite{dur360bev} and FTV360~\cite{maugey2019ftv360}. 
Dur360BEV provides outdoor driving 360$^\circ$ videos, but consists of a single moving-camera stream; we therefore construct a pseudo sparse multi-location setting by selecting frame pairs with approximately 10 m camera-center separation.
FTV360 provides calibrated multi-view 360$^\circ$ videos with intermediate viewpoints, but focuses on indoor scenes.
These two datasets provide complementary real-world evaluations beyond the controlled simulation benchmark.
\noindent\textbf{Quantitative results.}
Table~\ref{tab:real-world} shows the quantitative comparison between Stitch4D and baseline methods in the real-world experiments.
For each metric, the best result is highlighted in bold.
Stitch4D outperformed the baselines across all evaluation configurations.
These results show that Stitch4D achieves improved reconstruction quality for unobserved intermediate viewpoints on real-world datasets.

\begin{figure}[t]
    \centering
    \vspace{-1mm}
    \includegraphics[width=\linewidth]{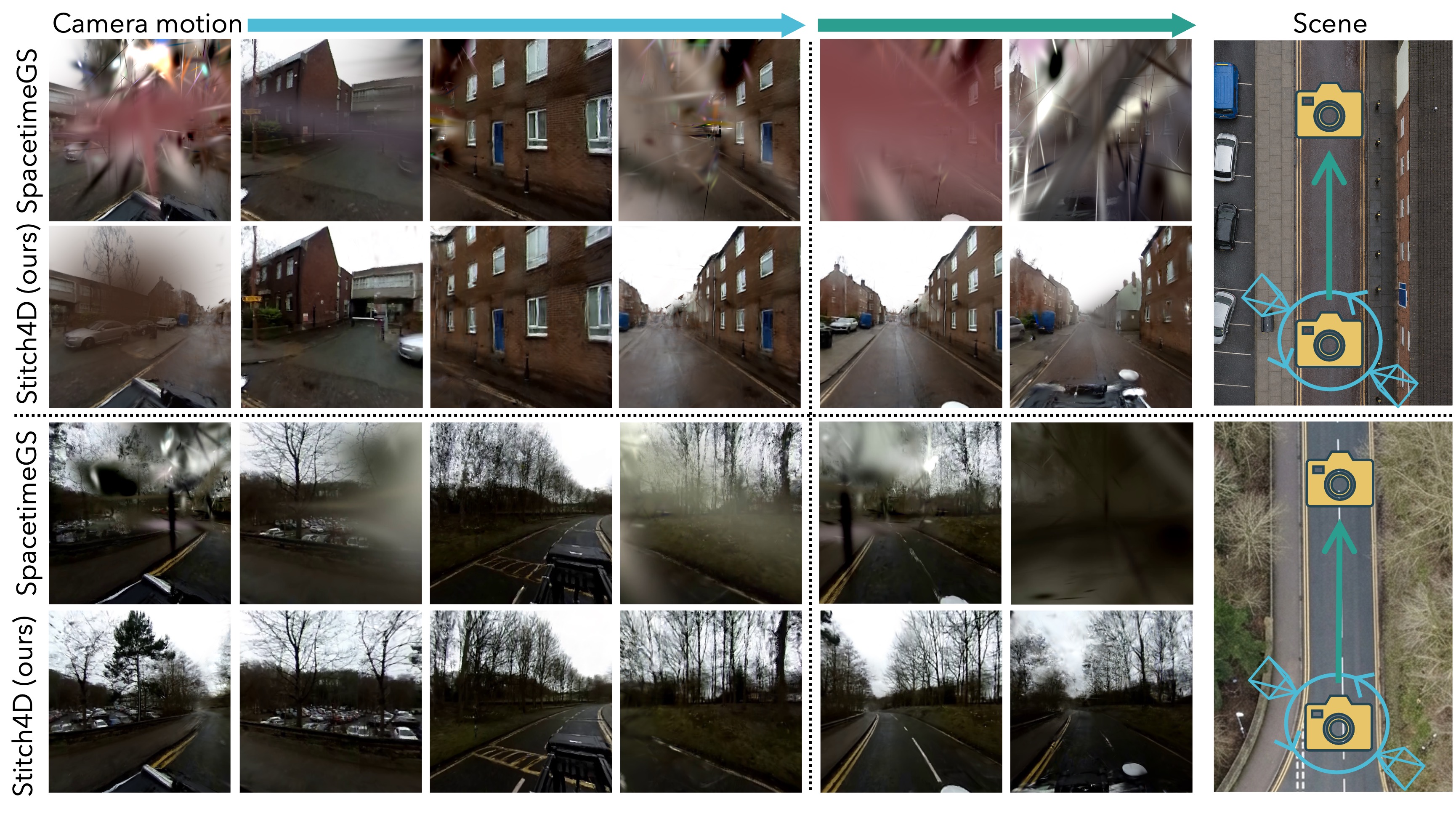}
    \vspace{-8mm}
    \caption{Real-world qualitative results on Dur360BEV.}
    \vspace{-7mm}
    \label{fig:real-world-success}
\end{figure}

\noindent\textbf{Qualitative results.}
Fig.~\ref{fig:real-world-success} shows qualitative results in real-world experiments on Dur360BEV~\cite{dur360bev}.
While the baselines produce discontinuous geometry, blurred structures, or inconsistent appearances, Stitch4D reconstructs coherent intermediate geometry and preserves scene appearance along the trajectory.
Further qualitative results on Dur360BEV and FTV360 are provided in the supplementary material.

\vspace{-4mm}
\section{Conclusion}
\vspace{-2mm}
We addressed the SP4DR problem, which aims to reconstruct a unified 4D representation from panoramic videos captured at spatially separated locations. 
To this end, we introduced Stitch4D, which combines MVBM for synthesizing intermediate viewpoints with MVJOM for jointly optimizing real and synthesized observations within a shared 4D representation. 
We also introduced U-S4D, a controlled benchmark with time-synchronized multi-location panoramic videos, calibrated camera poses, and ground-truth intermediate views for free-viewpoint evaluation. 
Experiments on U-S4D and two complementary real-world 360$^\circ$ video datasets demonstrated that Stitch4D consistently outperforms prior 4D reconstruction methods and achieves stable, globally consistent reconstruction under sparse multi-location observations. 
Future work will extend evaluation to fully synchronized real-world outdoor multi-location 360$^\circ$ capture settings and further investigate the integration of dynamic camera observations.

\clearpage

\subsubsection*{Acknowledgements.}
This work was partially supported by JST Moonshot and JST PRESTO (Grant Number JPMJPR22P8).

{
    \small
    \bibliographystyle{ieeenat_fullname}
    \bibliography{main}
}

\clearpage

\twocolumn[{
\begin{center}
\vspace{8mm}
{\LARGE\bfseries
Stitch4D: 4D Urban Reconstruction from Sparse Multi-Location Observations via Spatio-Temporal Interpolation
\par}
\vspace{5mm}
{\LARGE Supplementary Material\par}
\vspace{8mm}
\end{center}
}]

 

\setcounter{section}{0}
\setcounter{table}{0}
\setcounter{figure}{0}
\renewcommand{\thesection}{\Alph{section}}
\renewcommand{\thetable}{\Alph{table}}
\renewcommand{\thefigure}{\Alph{figure}}





\renewcommand{\thesection}{\Alph{section}}
\renewcommand{\thetable}{\Alph{table}}
\renewcommand{\thefigure}{\Alph{figure}}


%







\section{Method Details}

\subsection{Input Preparation}
For each input panoramic video $\bm{v}_i$, we estimate the depth for all frames.
We first compute panoramic optical flow between adjacent frames and derive dynamic masks from correspondences that violate motion consistency.
These masks suppress the influence of independently moving objects during depth estimation.
We then estimate depth for each frame conditioned on these dynamic masks using Marigold~\cite{ke2025marigold}.
This procedure improves temporal consistency and produces depth maps that are consistent across the entire panoramic sequence.

\begin{figure}[th]
    \centering
    \vspace{-2mm}
    \includegraphics[width=0.8\linewidth]{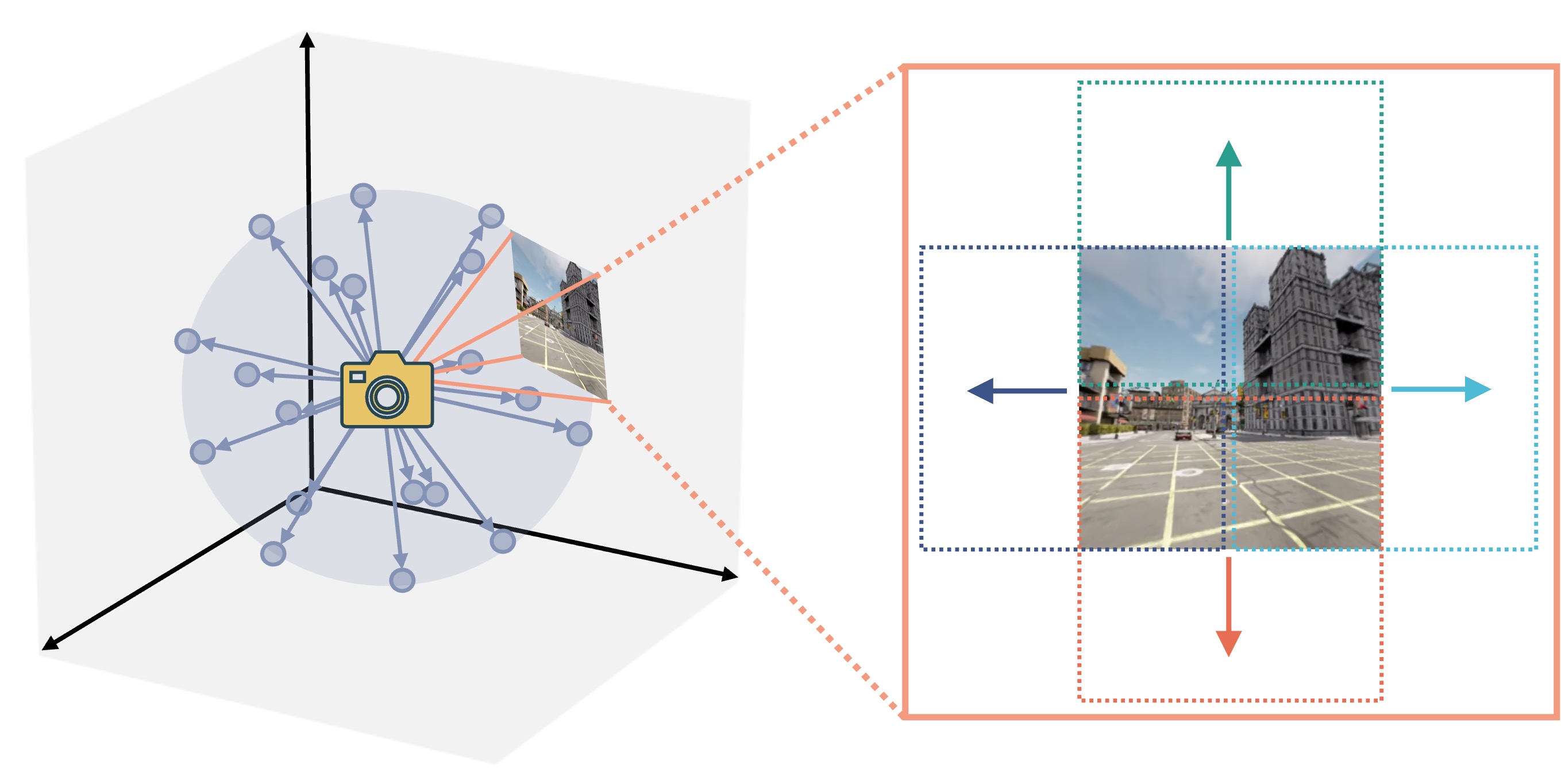}
    \vspace{-2mm}
    \caption{
    \textbf{Construction of 120 virtual perspective views from a panoramic video.}
(Left) Twenty uniformly distributed directions on the unit sphere.
(Right) For each direction, five views are generated using small angular offsets.
An additional 20 wide-FoV cameras provide global context.}
    \label{fig:5_4}
    \vspace{-3mm}
\end{figure}
In parallel, we convert each panoramic frame into perspective views for downstream reconstruction.
We adopt the standard cubemap representation and define the set of cubemaps as
$\mathcal{C} = \{\bm{C}_i \mid i \in \mathcal{N}_v\}$, where $\mathcal{N}_v$ denotes the set of camera locations.
Each cubemap $\bm{C}_i$ consists of six directional perspective videos,
$\bm{C}_i = \{ \bm{c}_i^d \mid d \in \mathcal{D} \},$
where $\mathcal{D} = \{\text{front}, \text{back}, \text{left}, \text{right}, \text{up}, \text{down}\}$ and $\bm{c}_i^d$ denotes the perspective video corresponding to direction $d$ at camera location $\bm{p}_i$.

To further improve spatial coverage, we generate a denser set of virtual perspective views.
We place virtual cameras on the unit sphere in 20 approximately uniformly distributed directions (Fig.~\ref{fig:5_4}, left).
For each direction, we define a central view and augment it with four additional views obtained by applying small angular offsets along the horizontal and vertical axes.
This yields five views per direction and 100 views in total (Fig.~\ref{fig:5_4}, right).
In addition, we introduce 20 wide field-of-view virtual cameras to capture broader spatial context that narrow field-of-view cameras may miss.
Consequently, each panoramic video is converted into a set of perspective views
$\mathcal{U}_i = \{ \bm{u}_i^{(v)} \mid v \in \mathcal{V} \}$, $|\mathcal{V}| = 120$.
These sets provide dense multi-directional observations that better capture spatial structure and temporal changes in dynamic urban scenes.

\section{Multi-Video Joint Optimization Module}

\begin{figure}[!t]
    \centering
    \includegraphics[width=\linewidth]{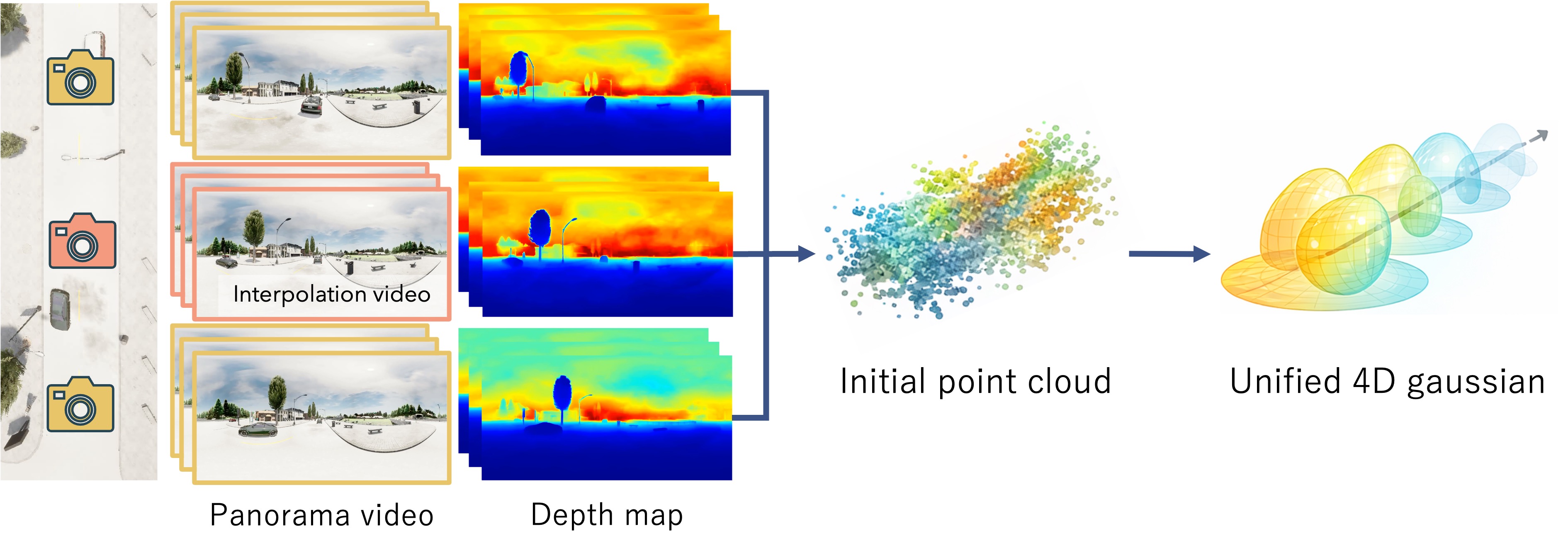}
    \caption{
     \textbf{Structural overview of MVJOM.}
     MVJOM jointly optimizes panoramic videos from multiple camera locations to reconstruct a unified time-varying scene representation.
    }
    \label{fig:mvjom}
    \vspace{-3mm}
\end{figure}

As shown in Fig.~\ref{fig:mvjom}, MVJOM integrates panoramic videos captured from multiple camera locations and jointly optimizes a single time-varying scene representation $\mathcal{G}(t)$ in a unified coordinate frame. 
This joint optimization enables the model to enforce spatial consistency across locations throughout the sequence, which is difficult to achieve when each location is optimized independently.

\section{Seam-Aware Inter-Location Loss}





This section provides the detailed definitions of the reconstruction loss and the seam-aware weighting functions.

The reconstruction loss $\mathcal{L}_{\mathrm{recon}}$ is defined as
\begin{equation*}
\resizebox{1\linewidth}{!}{$
\mathcal{L}_{\mathrm{recon}}
=
(1-\lambda_{\mathrm{dssim}})
\|\hat{\mathbf{I}} - \mathbf{I}\|_1
+
\lambda_{\mathrm{dssim}}
\bigl(1-\mathrm{SSIM}(\hat{\mathbf{I}}, \mathbf{I})\bigr)
+
\lambda_{\mathrm{reg}}
\mathcal{L}_{\mathrm{reg}},
$}
\end{equation*}
where $\hat{\mathbf{I}}$ and $\mathbf{I}$ denote the rendered and ground-truth RGB images, respectively.
This loss combines an $\ell_1$ photometric term, an SSIM term, and an additional regularization term.
The weights $\lambda_{\mathrm{dssim}}$ and $\lambda_{\mathrm{reg}}$ control the contributions of the SSIM and regularization terms, respectively.

The distance-aware seam weight $\beta(\delta)$ is defined as
\begin{equation*}
\beta(\delta)
=
1+
\lambda_{\mathrm{seam}}
\exp\!\left(
-\frac{\delta^2}{2\tau^2}
\right),
\label{eq:beta}
\end{equation*}
where $\delta$ is the distance to the nearest camera-location boundary.
The parameter $\lambda_{\mathrm{seam}}$ controls the strength of seam emphasis, while $\tau$ determines the spatial extent of the boundary region.

The boundary-aware inter-location weight $\gamma(\delta)$ is defined as
\begin{equation*}
\gamma(\delta)
=
\exp\!\left(
-\frac{\delta^2}{2\tau_{\mathrm{inter}}^2}
\right),
\label{eq:gamma}
\end{equation*}
where $\tau_{\mathrm{inter}}$ controls the spatial extent over which the inter-location consistency term is applied.
Unlike $\beta(\delta)$, which increases the reconstruction loss near boundaries, $\gamma(\delta)$ acts as a local mask that suppresses inter-location constraints away from connection regions.

The inter-location term is applied only to neighboring viewpoint pairs sampled near camera-location boundaries.
This local application keeps the constraint focused on transition regions and avoids imposing dense correspondences across distant or weakly overlapping views.

\section{Experimental Setup Details}

\subsection{Implementation Details}
\begin{table}[t]
    \setlength{\tabcolsep}{4pt}
    \renewcommand*{\arraystretch}{1.25}
    \newcommand*{\bhline}[1]{\noalign{\hrule height #1}}
    \caption{Experimental setup for \proposed.}
    \centering
    \begin{tabular}{@{\hspace{5pt}}l@{\hspace{30pt}}l@{\hspace{5pt}}}
        \bhline{1.0pt}
        Optimizer & Adam ($\beta_1=0.9, \beta_2=0.999$) \\
        Feature learning rate & $0.001$ \\
        Opacity learning rate & $0.05$ \\
        Scaling learning rate & $0.005$ \\
        Rotation learning rate & $0.001$ \\
        Iterations & $60000$ \\
        Batch size & $4$ \\
        \bhline{1.0pt}
    \end{tabular}
    \label{tab:exp_setup}
    \vspace{-3mm}
\end{table}
Table~\ref{tab:exp_setup} summarizes the optimization settings used in our experiments.

\subsection{Metrics}
PSNR is defined as follows:
\begin{align*}
\text{PSNR} = 10 \log_{10} \left( \frac{\mathrm{MAX}_\mathbf{I}^2}{\text{MSE}} \right).
\end{align*}
Here, MSE denotes the mean squared error, which is defined as follows:
\begin{align*}
\text{MSE} = \frac{1}{HW} \sum_{h=1}^{H} \sum_{w=1}^{W} \left( \mathbf{I}(h,w) - \hat{\mathbf{I}}(h,w) \right)^2,
\end{align*}
where $\mathrm{MAX}_\mathbf{I}$ denotes the maximum pixel value.
SSIM~\cite{ssim} is defined as follows:
\begin{align*}
\mathrm{SSIM}(\mathbf{I}, \hat{\mathbf{I}}) =
\frac{(2\mu_\mathbf{I} \mu_{\hat{\mathbf{I}}} + C_1)(2\sigma_{\mathrm{loc},\mathbf{I},\hat{\mathbf{I}}} + C_2)}
{(\mu_\mathbf{I}^2 + \mu_{\hat{\mathbf{I}}}^2 + C_1)(\sigma_{\mathrm{loc},\mathbf{I}}^2 + \sigma_{\mathrm{loc},\hat{\mathbf{I}}}^2 + C_2)},
\end{align*}
where $\mu_\mathbf{I}$, $\sigma_{\mathrm{loc},\hat{\mathbf{I}}}$, and $\sigma_{\mathrm{loc},\mathbf{I},\hat{\mathbf{I}}}$ denote the mean intensities of image $\mathbf{I}$, its variance, and the covariance between $\mathbf{I}$ and $\hat{\mathbf{I}}$, respectively.
Constants $C_1$ and $C_2$ are introduced for numerical stability.
LPIPS~\cite{Zhang_2018_CVPR_lpips} is defined as follows:
\begin{align*}
\resizebox{0.9\linewidth}{!}{$
\mathrm{LPIPS}(\mathbf{I}, \hat{\mathbf{I}}) 
=
\sum_{l}
\frac{1}{U_l V_l}
\sum_{p=1}^{U_l} \sum_{q=1}^{V_l}
\left\|
\mathbf{w}_l \odot
\left(
\phi_l(\mathbf{I})_{p,q}
-
\phi_l(\hat{\mathbf{I}})_{p,q}
\right)
\right\|_2^2 .
$}
\end{align*}
where $\phi_l(\cdot)$ and $\phi_l(\mathbf{I})_{p,q}$ denote the feature extractor of a pretrained CNN at layer $l$ and the feature vector at spatial index $(p,q)$ on the $U_l \times V_l$ feature map, respectively. 
Moreover, $\mathbf{w}_l$, $\odot$, and $\|\cdot\|_2$ denote a learned channel-wise weight vector at layer $l$, element-wise multiplication, and the $\ell_2$ norm, respectively.

\section{Additional Results}

\subsection{Quantitative Results}


\begin{table*}[t]
\centering
\caption{Quantitative comparison of Stitch4D built on different 4D representation backbones and baseline methods under the full reconstruction and temporal split settings. The best score for each metric is shown in \textbf{bold}.}
\vspace{-1mm}
\resizebox{\textwidth}{!}{%
\begin{tabular}{lcccccccccccc}
\toprule
\multirow{3}{*}{Method}
& \multicolumn{6}{c}{Full reconstruction}
& \multicolumn{6}{c}{Temporal split} \\
\cmidrule(lr){2-7}
\cmidrule(lr){8-13}
& \multicolumn{3}{c}{Trajectory interpolation}
& \multicolumn{3}{c}{Seen-viewpoints}
& \multicolumn{3}{c}{Trajectory interpolation}
& \multicolumn{3}{c}{Seen-viewpoints} \\
\cmidrule(lr){2-4}
\cmidrule(lr){5-7}
\cmidrule(lr){8-10}
\cmidrule(lr){11-13}
& \small PSNR $\uparrow$ & \small SSIM $\uparrow$ & \small LPIPS $\downarrow$
& \small PSNR $\uparrow$ & \small SSIM $\uparrow$ & \small LPIPS $\downarrow$
& \small PSNR $\uparrow$ & \small SSIM $\uparrow$ & \small LPIPS $\downarrow$
& \small PSNR $\uparrow$ & \small SSIM $\uparrow$ & \small LPIPS $\downarrow$ \\
\midrule
4DGS~\cite{wu20244dgs}
& 11.51 & 0.28 & 0.84
& 15.79 & 0.58 & 0.84
& 10.54 & 0.25 & 0.80
& 13.78 & 0.52 & 0.64 \\
\makecell[tl]{SpacetimeGS \cite{li2024spacetime}}
& 12.91 & 0.55 & 0.67
& 17.75 & 0.79 & 0.33
& 12.47 & 0.54 & 0.69
& 17.49 & 0.78 & 0.33 \\

\makecell[tl]{FreeTimeGS \cite{wang2025freetimegs}}
& 11.75 & 0.52 & 0.75
& 16.70 & 0.71 & 0.41
& 11.84 & 0.53 & 0.74
& 16.53 & 0.71 & 0.41 \\
\midrule
Stitch4D (SpacetimeGS)
& 15.31 & \textbf{0.60} & 0.51
& \textbf{26.34} & \textbf{0.92} & \textbf{0.13}
& 14.88 & \textbf{0.59} & 0.52
& \textbf{24.63} & \textbf{0.90} & \textbf{0.15} \\
Stitch4D (FreeTimeGS)
& \textbf{15.98} & \textbf{0.59} & \textbf{0.50}
& 24.56 & 0.88 & 0.18
& \textbf{16.01} & \textbf{0.59} & \textbf{0.50}
& 23.13 & 0.85 & 0.20 \\
\bottomrule
\end{tabular}%
}
\vspace{-2mm}
\label{tab:quantitative_backbone}
\end{table*}

\begin{table*}[t]
\centering
\caption{Quantitative results of the ablation study on Stitch4D built on FreeTimeGS under the full reconstruction and temporal split settings. The best score for each metric is shown in \textbf{bold}.}
\vspace{-1mm}
\resizebox{\textwidth}{!}{%
\begin{tabular}{lcccccccccccc}
\toprule
\multirow{3}{*}{Method}
& \multicolumn{6}{c}{Full reconstruction}
& \multicolumn{6}{c}{Temporal split} \\
\cmidrule(lr){2-7}
\cmidrule(lr){8-13}
& \multicolumn{3}{c}{Trajectory interpolation}
& \multicolumn{3}{c}{Seen-viewpoints}
& \multicolumn{3}{c}{Trajectory interpolation}
& \multicolumn{3}{c}{Seen-viewpoints} \\
\cmidrule(lr){2-4}
\cmidrule(lr){5-7}
\cmidrule(lr){8-10}
\cmidrule(lr){11-13}
& \small PSNR $\uparrow$ & \small SSIM $\uparrow$ & \small LPIPS $\downarrow$
& \small PSNR $\uparrow$ & \small SSIM $\uparrow$ & \small LPIPS $\downarrow$
& \small PSNR $\uparrow$ & \small SSIM $\uparrow$ & \small LPIPS $\downarrow$
& \small PSNR $\uparrow$ & \small SSIM $\uparrow$ & \small LPIPS $\downarrow$ \\
\midrule
Stitch4D (FreeTimeGS) 
& \textbf{15.98} & \textbf{0.59} & \textbf{0.50}
& \textbf{24.56} & \textbf{0.88} & \textbf{0.18}
& \textbf{16.01} & \textbf{0.59} & \textbf{0.50}
& \textbf{23.13} & \textbf{0.85} & \textbf{0.20} \\
w/o MVBM
& 14.81 & 0.58 & 0.62
& 23.76 & 0.85 & 0.23
& 14.90 & 0.58 & 0.62
& 22.24 & 0.82 & 0.26 \\
w/o MVJOM, MVBM
& 11.75 & 0.52 & 0.75
& 16.70 & 0.71 & 0.41
& 11.84 & 0.53 & 0.74
& 16.53 & 0.71 & 0.41 \\
\bottomrule
\end{tabular}%
}
\vspace{-2mm}
\label{tab:freetime_comparison}
\end{table*}
To examine whether the effectiveness of Stitch4D depends on the choice of 4D representation backbone, we further replace the original SpacetimeGS backbone with FreeTimeGS and compare the resulting variants under the full reconstruction and temporal split settings.

Table~\ref{tab:quantitative_backbone} shows the quantitative comparison of Stitch4D built on different 4D representation backbones and baseline methods under the full reconstruction and temporal split settings.
Across all settings, Stitch4D consistently outperforms the baseline methods, while showing comparable performance across different backbones.
These results suggest that the effectiveness of Stitch4D is not specific to a particular 4D representation backbone.

Moreover, Table~\ref{tab:freetime_comparison} shows the quantitative results of ablation studies for the FreeTimeGS-based variant of Stitch4D under the full reconstruction and temporal split settings.
Removing MVBM degrades the performance across all settings, and further removing MVJOM leads to an even larger drop.
These results indicate that both MVBM and MVJOM contribute to the performance of Stitch4D even in different 4D representation backbones.

Furthermore, Tables~\ref{tab:quantitative_urban1}, \ref{tab:quantitative_urban2}, and \ref{tab:quantitative_urban3} report per-scene quantitative results on the U-S4D benchmark under the full reconstruction and temporal split settings.
\begin{table*}[t!]
\centering
\caption{
Quantitative comparison with baseline methods on Scene 1 and Scene 2 of Urban Area 1 in the full reconstruction and temporal split settings.
}
\vspace{-1mm}
\setlength{\tabcolsep}{2.2pt}
\renewcommand{\arraystretch}{1.05}
\resizebox{\textwidth}{!}{%
\begin{tabular}{@{}l*{24}{c}@{}}
\toprule
\multirow{4}{*}{Method}
& \multicolumn{12}{c}{Scene 1}
& \multicolumn{12}{c}{Scene 2} \\
\cmidrule(lr){2-13}
\cmidrule(lr){14-25}

& \multicolumn{6}{c}{Full reconstruction}
& \multicolumn{6}{c}{Temporal split}
& \multicolumn{6}{c}{Full reconstruction}
& \multicolumn{6}{c}{Temporal split} \\
\cmidrule(lr){2-7}
\cmidrule(lr){8-13}
\cmidrule(lr){14-19}
\cmidrule(lr){20-25}

& \multicolumn{3}{c}{Trajectory interpolation}
& \multicolumn{3}{c}{Seen-viewpoints}
& \multicolumn{3}{c}{Trajectory interpolation}
& \multicolumn{3}{c}{Seen-viewpoints}
& \multicolumn{3}{c}{Trajectory interpolation}
& \multicolumn{3}{c}{Seen-viewpoints}
& \multicolumn{3}{c}{Trajectory interpolation}
& \multicolumn{3}{c}{Seen-viewpoints} \\
\cmidrule(lr){2-4}
\cmidrule(lr){5-7}
\cmidrule(lr){8-10}
\cmidrule(lr){11-13}
\cmidrule(lr){14-16}
\cmidrule(lr){17-19}
\cmidrule(lr){20-22}
\cmidrule(lr){23-25}

& \scriptsize PSNR $\uparrow$
& \scriptsize SSIM $\uparrow$
& \scriptsize LPIPS $\downarrow$
& \scriptsize PSNR $\uparrow$
& \scriptsize SSIM $\uparrow$
& \scriptsize LPIPS $\downarrow$
& \scriptsize PSNR $\uparrow$
& \scriptsize SSIM $\uparrow$
& \scriptsize LPIPS $\downarrow$
& \scriptsize PSNR $\uparrow$
& \scriptsize SSIM $\uparrow$
& \scriptsize LPIPS $\downarrow$
& \scriptsize PSNR $\uparrow$
& \scriptsize SSIM $\uparrow$
& \scriptsize LPIPS $\downarrow$
& \scriptsize PSNR $\uparrow$
& \scriptsize SSIM $\uparrow$
& \scriptsize LPIPS $\downarrow$
& \scriptsize PSNR $\uparrow$
& \scriptsize SSIM $\uparrow$
& \scriptsize LPIPS $\downarrow$
& \scriptsize PSNR $\uparrow$
& \scriptsize SSIM $\uparrow$
& \scriptsize LPIPS $\downarrow$ \\
\midrule

\rowcolor{gray!12}
\multicolumn{25}{l}{\textbf{Urban scene reconstruction methods}} \\

\makecell[tl]{PVG \cite{chen2026periodic}}
& 12.48 & 0.40 & 0.95
& 12.77 & 0.53 & 0.70
& 11.86 & 0.39 & 0.99
& 12.56 & 0.52 & 0.72

& 12.92 & 0.36 & 0.90
& 13.31 & 0.52 & 0.75
& 12.78 & 0.36 & 0.91
& 12.89 & 0.51 & 0.77 \\

\makecell[tl]{Street Gaussians \cite{yan2024street}}
& 11.31 & 0.34 & 0.89
& 17.36 & 0.62 & 0.65
& 10.42 & 0.26 & 0.83
& 18.20 & 0.63 & 0.63

& 11.28 & 0.27 & 0.83
& 14.13 & 0.51 & 0.76
& 11.26 & 0.22 & 0.75
& 15.81 & 0.55 & 0.70 \\

\rowcolor{gray!12}
\multicolumn{25}{l}{\textbf{General 4D reconstruction methods}} \\

\makecell[tl]{SpacetimeGS \cite{li2024spacetime}}
& 13.29 & 0.35 & 0.73
& 17.16 & 0.73 & 0.34
& 12.88 & 0.35 & 0.75
& 16.93 & 0.73 & 0.34

& 12.97 & 0.36 & 0.77
& 17.64 & 0.76 & 0.28
& 12.72 & 0.36 & 0.77
& 16.31 & 0.74 & 0.30 \\

\makecell[tl]{FreeTimeGS \cite{wang2025freetimegs}}
& 10.77 & 0.26 & 0.87
& 13.15 & 0.33 & 0.72
& 10.90 & 0.25 & 0.87
& 11.48 & 0.40 & 0.73

& 11.36 & 0.32 & 0.92
& 17.28 & 0.74 & 0.34
& 11.20 & 0.32 & 0.92
& 15.80 & 0.67 & 0.39 \\

\midrule

\makecell[tl]{\proposed~(ours)}
& 16.34 & 0.44 & 0.52
& 22.90 & 0.87 & 0.18
& 16.41 & 0.44 & 0.52
& 20.98 & 0.84 & 0.21

& 16.05 & 0.43 & 0.59
& 21.90 & 0.88 & 0.17
& 15.00 & 0.39 & 0.61
& 21.02 & 0.87 & 0.17 \\

\bottomrule
\end{tabular}%
}
\label{tab:quantitative_urban1}
\vspace{-2mm}
\end{table*}

\begin{table*}[t!]
\centering
\caption{
Quantitative comparison with baseline methods on Scene 1 and Scene 2 of Urban Area 2 in the full reconstruction and temporal split settings.
}
\vspace{-1mm}
\setlength{\tabcolsep}{2.2pt}
\renewcommand{\arraystretch}{1.05}
\resizebox{\textwidth}{!}{%
\begin{tabular}{@{}l*{24}{c}@{}}
\toprule
\multirow{4}{*}{Method}
& \multicolumn{12}{c}{Scene 1}
& \multicolumn{12}{c}{Scene 2} \\
\cmidrule(lr){2-13}
\cmidrule(lr){14-25}

& \multicolumn{6}{c}{Full reconstruction}
& \multicolumn{6}{c}{Temporal split}
& \multicolumn{6}{c}{Full reconstruction}
& \multicolumn{6}{c}{Temporal split} \\
\cmidrule(lr){2-7}
\cmidrule(lr){8-13}
\cmidrule(lr){14-19}
\cmidrule(lr){20-25}

& \multicolumn{3}{c}{Trajectory interpolation}
& \multicolumn{3}{c}{Seen-viewpoints}
& \multicolumn{3}{c}{Trajectory interpolation}
& \multicolumn{3}{c}{Seen-viewpoints}
& \multicolumn{3}{c}{Trajectory interpolation}
& \multicolumn{3}{c}{Seen-viewpoints}
& \multicolumn{3}{c}{Trajectory interpolation}
& \multicolumn{3}{c}{Seen-viewpoints} \\
\cmidrule(lr){2-4}
\cmidrule(lr){5-7}
\cmidrule(lr){8-10}
\cmidrule(lr){11-13}
\cmidrule(lr){14-16}
\cmidrule(lr){17-19}
\cmidrule(lr){20-22}
\cmidrule(lr){23-25}

& \scriptsize PSNR $\uparrow$
& \scriptsize SSIM $\uparrow$
& \scriptsize LPIPS $\downarrow$
& \scriptsize PSNR $\uparrow$
& \scriptsize SSIM $\uparrow$
& \scriptsize LPIPS $\downarrow$
& \scriptsize PSNR $\uparrow$
& \scriptsize SSIM $\uparrow$
& \scriptsize LPIPS $\downarrow$
& \scriptsize PSNR $\uparrow$
& \scriptsize SSIM $\uparrow$
& \scriptsize LPIPS $\downarrow$
& \scriptsize PSNR $\uparrow$
& \scriptsize SSIM $\uparrow$
& \scriptsize LPIPS $\downarrow$
& \scriptsize PSNR $\uparrow$
& \scriptsize SSIM $\uparrow$
& \scriptsize LPIPS $\downarrow$
& \scriptsize PSNR $\uparrow$
& \scriptsize SSIM $\uparrow$
& \scriptsize LPIPS $\downarrow$
& \scriptsize PSNR $\uparrow$
& \scriptsize SSIM $\uparrow$
& \scriptsize LPIPS $\downarrow$ \\
\midrule

\rowcolor{gray!12}
\multicolumn{25}{l}{\textbf{Urban scene reconstruction methods}} \\

\makecell[tl]{PVG \cite{chen2026periodic}}
& 14.28 & 0.72 & 0.67
& 14.29 & 0.81 & 0.48
& 14.28 & 0.72 & 0.68
& 14.85 & 0.81 & 0.48

& 12.04 & 0.71 & 0.67
& 14.64 & 0.80 & 0.44
& 12.18 & 0.68 & 0.66
& 14.63 & 0.80 & 0.45 \\

\makecell[tl]{Street Gaussians \cite{yan2024street}}
& 13.23 & 0.70 & 0.70
& 16.12 & 0.81 & 0.47
& 12.99 & 0.71 & 0.71
& 15.86 & 0.80 & 0.48

& 11.31 & 0.71 & 0.70
& 16.96 & 0.83 & 0.39
& 10.08 & 0.69 & 0.75
& 17.12 & 0.82 & 0.41 \\

\rowcolor{gray!12}
\multicolumn{25}{l}{\textbf{General 4D reconstruction methods}} \\

\makecell[tl]{SpacetimeGS \cite{li2024spacetime}}
& 14.98 & 0.71 & 0.59
& 18.52 & 0.82 & 0.36
& 14.48 & 0.70 & 0.61
& 17.99 & 0.78 & 0.39

& 11.21 & 0.60 & 0.69
& 16.69 & 0.79 & 0.35
& 9.72 & 0.59 & 0.72
& 17.83 & 0.83 & 0.31 \\

\makecell[tl]{FreeTimeGS \cite{wang2025freetimegs}}
& 13.25 & 0.67 & 0.67
& 16.38 & 0.83 & 0.39
& 13.33 & 0.68 & 0.69
& 17.32 & 0.74 & 0.39

& 11.01 & 0.60 & 0.67
& 16.32 & 0.72 & 0.37
& 11.30 & 0.65 & 0.65
& 18.08 & 0.84 & 0.31 \\

\midrule

\makecell[tl]{\proposed~(ours)}
& 17.53 & 0.73 & 0.42
& 29.50 & 0.96 & 0.09
& 17.05 & 0.72 & 0.44
& 28.26 & 0.95 & 0.10

& 12.81 & 0.64 & 0.58
& 29.85 & 0.96 & 0.09
& 11.65 & 0.63 & 0.60
& 28.48 & 0.95 & 0.10 \\

\bottomrule
\end{tabular}%
}
\vspace{-2mm}
\label{tab:quantitative_urban2}
\end{table*}
\begin{table*}[t!]
\centering
\caption{
Quantitative comparison with baseline methods on Scene 1 and Scene 2 of Urban Area 3 in the full reconstruction and temporal split settings.
}
\vspace{-1mm}
\setlength{\tabcolsep}{2.2pt}
\renewcommand{\arraystretch}{1.05}
\resizebox{\textwidth}{!}{%
\begin{tabular}{@{}l*{24}{c}@{}}
\toprule
\multirow{4}{*}{Method}
& \multicolumn{12}{c}{Scene 1}
& \multicolumn{12}{c}{Scene 2} \\
\cmidrule(lr){2-13}
\cmidrule(lr){14-25}

& \multicolumn{6}{c}{Full reconstruction}
& \multicolumn{6}{c}{Temporal split}
& \multicolumn{6}{c}{Full reconstruction}
& \multicolumn{6}{c}{Temporal split} \\
\cmidrule(lr){2-7}
\cmidrule(lr){8-13}
\cmidrule(lr){14-19}
\cmidrule(lr){20-25}

& \multicolumn{3}{c}{Trajectory interpolation}
& \multicolumn{3}{c}{Seen-viewpoints}
& \multicolumn{3}{c}{Trajectory interpolation}
& \multicolumn{3}{c}{Seen-viewpoints}
& \multicolumn{3}{c}{Trajectory interpolation}
& \multicolumn{3}{c}{Seen-viewpoints}
& \multicolumn{3}{c}{Trajectory interpolation}
& \multicolumn{3}{c}{Seen-viewpoints} \\
\cmidrule(lr){2-4}
\cmidrule(lr){5-7}
\cmidrule(lr){8-10}
\cmidrule(lr){11-13}
\cmidrule(lr){14-16}
\cmidrule(lr){17-19}
\cmidrule(lr){20-22}
\cmidrule(lr){23-25}

& \scriptsize PSNR $\uparrow$
& \scriptsize SSIM $\uparrow$
& \scriptsize LPIPS $\downarrow$
& \scriptsize PSNR $\uparrow$
& \scriptsize SSIM $\uparrow$
& \scriptsize LPIPS $\downarrow$
& \scriptsize PSNR $\uparrow$
& \scriptsize SSIM $\uparrow$
& \scriptsize LPIPS $\downarrow$
& \scriptsize PSNR $\uparrow$
& \scriptsize SSIM $\uparrow$
& \scriptsize LPIPS $\downarrow$
& \scriptsize PSNR $\uparrow$
& \scriptsize SSIM $\uparrow$
& \scriptsize LPIPS $\downarrow$
& \scriptsize PSNR $\uparrow$
& \scriptsize SSIM $\uparrow$
& \scriptsize LPIPS $\downarrow$
& \scriptsize PSNR $\uparrow$
& \scriptsize SSIM $\uparrow$
& \scriptsize LPIPS $\downarrow$
& \scriptsize PSNR $\uparrow$
& \scriptsize SSIM $\uparrow$
& \scriptsize LPIPS $\downarrow$ \\
\midrule

\rowcolor{gray!12}
\multicolumn{25}{l}{\textbf{Urban scene reconstruction methods}} \\

\makecell[tl]{PVG \cite{chen2026periodic}}
& 12.45 & 0.56 & 0.74
& 12.54 & 0.72 & 0.56
& 12.89 & 0.63 & 0.62
& 15.05 & 0.78 & 0.49

& 12.87 & 0.73 & 0.61
& 14.99 & 0.78 & 0.49
& 12.00 & 0.64 & 0.75
& 12.86 & 0.73 & 0.55 \\

\makecell[tl]{Street Gaussians \cite{yan2024street}}
& 11.95 & 0.59 & 0.76
& 16.50 & 0.75 & 0.48
& 11.53 & 0.59 & 0.77
& 19.39 & 0.78 & 0.42

& 12.00 & 0.72 & 0.62
& 16.03 & 0.78 & 0.48
& 12.67 & 0.72 & 0.66
& 17.93 & 0.80 & 0.41 \\

\rowcolor{gray!12}
\multicolumn{25}{l}{\textbf{General 4D reconstruction methods}} \\

\makecell[tl]{SpacetimeGS \cite{li2024spacetime}}
& 12.61 & 0.66 & 0.62
& 17.17 & 0.80 & 0.38
& 12.20 & 0.65 & 0.64
& 16.94 & 0.80 & 0.38

& 12.43 & 0.60 & 0.65
& 19.34 & 0.84 & 0.26
& 12.81 & 0.60 & 0.64
& 18.93 & 0.82 & 0.28 \\

\makecell[tl]{FreeTimeGS \cite{wang2025freetimegs}}
& 12.07 & 0.66 & 0.65
& 17.16 & 0.81 & 0.37
& 12.28 & 0.67 & 0.64
& 17.36 & 0.81 & 0.37

& 12.05 & 0.59 & 0.71
& 19.90 & 0.85 & 0.27
& 12.00 & 0.60 & 0.70
& 19.15 & 0.82 & 0.29 \\

\midrule

\makecell[tl]{\proposed~(ours)}
& 14.06 & 0.69 & 0.47
& 27.20 & 0.94 & 0.11
& 14.11 & 0.69 & 0.47
& 23.84 & 0.90 & 0.15

& 15.08 & 0.64 & 0.49
& 26.68 & 0.93 & 0.13
& 15.07 & 0.65 & 0.50
& 25.18 & 0.91 & 0.14 \\

\bottomrule
\end{tabular}%
}
\vspace{-2mm}
\label{tab:quantitative_urban3}
\end{table*}

\subsection{Ablation Study}

\begin{table}[t!]
\centering
\caption{Additional quantitative results of the ablation studies under the temporal split setting.
    The best values for each metric are highlighted in \textbf{bold}.}
\vspace{-2mm}
\resizebox{\columnwidth}{!}
{
\begin{tabular}{lcccccc}
\toprule
\multirow{2}{*}{Method}
& \multicolumn{3}{c}{Trajectory interpolation}
& \multicolumn{3}{c}{Seen-viewpoints}\\
\cmidrule(lr){2-4}
\cmidrule(lr){5-7}
& \small PSNR $\uparrow$
& \small SSIM $\uparrow$
& \small LPIPS $\downarrow$
& \small PSNR $\uparrow$
& \small SSIM $\uparrow$
& \small LPIPS $\downarrow$
\\
\midrule 

\makecell[tl]{w/o MVBM}
& \makecell[tc]{13.80}
& \makecell[tc]{0.54}
& \makecell[tc]{0.61}
& \makecell[tc]{23.10}
& \makecell[tc]{0.83}
& \makecell[tc]{0.20}
\\

\makecell[tl]{w/o MVJOM, MVBM}
& \makecell[tc]{12.47}
& \makecell[tc]{0.54}
& \makecell[tc]{0.69}
& \makecell[tc]{17.49}
& \makecell[tc]{0.78}
& \makecell[tc]{0.33}
\\

\makecell[tl]{Stitch4D (ours)}
& \makecell[tc]{\textbf{14.88}}
& \makecell[tc]{\textbf{0.59}}
& \makecell[tc]{\textbf{0.52}}
& \makecell[tc]{\textbf{24.63}}
& \makecell[tc]{\textbf{0.90}}
& \makecell[tc]{\textbf{0.15}}
\\

\bottomrule
\end{tabular}
}
\label{tab:ablation_2}
\end{table}
Table~\ref{tab:ablation_2} shows additional quantitative ablation results for Stitch4D (SpacetimeGS) in the temporal split setting. The best score for each metric is shown in \textbf{bold}.
Similar to the full reconstruction setting, comparisons between the full model and its ablated variants show that the model with all modules consistently achieves better performance, indicating that both MVBM and MVJOM contribute to the overall performance even in the temporal split setting.

\begin{table}[t!]
\centering
\caption{Additional quantitative results of the ablation study on the seam-aware inter-location loss under the full reconstruction setting.}
\vspace{-4mm}
\resizebox{\columnwidth}{!}
{
\begin{tabular}{lcccccc}
\toprule
\multirow{2}{*}{Model}
& \multicolumn{3}{c}{Trajectory interpolation}
& \multicolumn{3}{c}{Seen-viewpoints}\\
\cmidrule(lr){2-4}
\cmidrule(lr){5-7}
& \small PSNR  $\uparrow$
& \small SSIM $\uparrow$
& \small LPIPS $\downarrow$
& \small PSNR $\uparrow$
& \small SSIM $\uparrow$
& \small LPIPS $\downarrow$
\\
\midrule 

\makecell[tl]{w/o $\mathcal{L}_{\mathrm{inter}}$}
& \makecell[tc]{14.54}
& \makecell[tc]{0.57}
& \makecell[tc]{0.53}
& \makecell[tc]{\textbf{26.44}}
& \makecell[tc]{\textbf{0.92}}
& \makecell[tc]{\textbf{0.13}}
\\

\makecell[tl]{Stitch4D (ours)}
& \makecell[tc]{\textbf{15.31}}
& \makecell[tc]{\textbf{0.60}}
& \makecell[tc]{\textbf{0.51}}
& \makecell[tc]{26.34}
& \makecell[tc]{\textbf{0.92}}
& \makecell[tc]{\textbf{0.13}}
\\

\bottomrule
\end{tabular}
}
\vspace{-3mm}
\label{tab:loss_ablation}
\end{table}

Table~\ref{tab:loss_ablation} shows the ablation of the seam-aware inter-location loss.
While the seen-viewpoint metrics remain comparable, adding the loss improves trajectory interpolation performance.
This suggests that the loss contributes to reconstruction quality in unobserved regions between endpoint observations, particularly around inter-location seams.

\subsection{Qualitative Results}

\subsubsection{Additional Real-World Qualitative Results.}
\begin{figure*}[t]
    \centering
    \vspace{-1mm}
    \includegraphics[width=\textwidth]{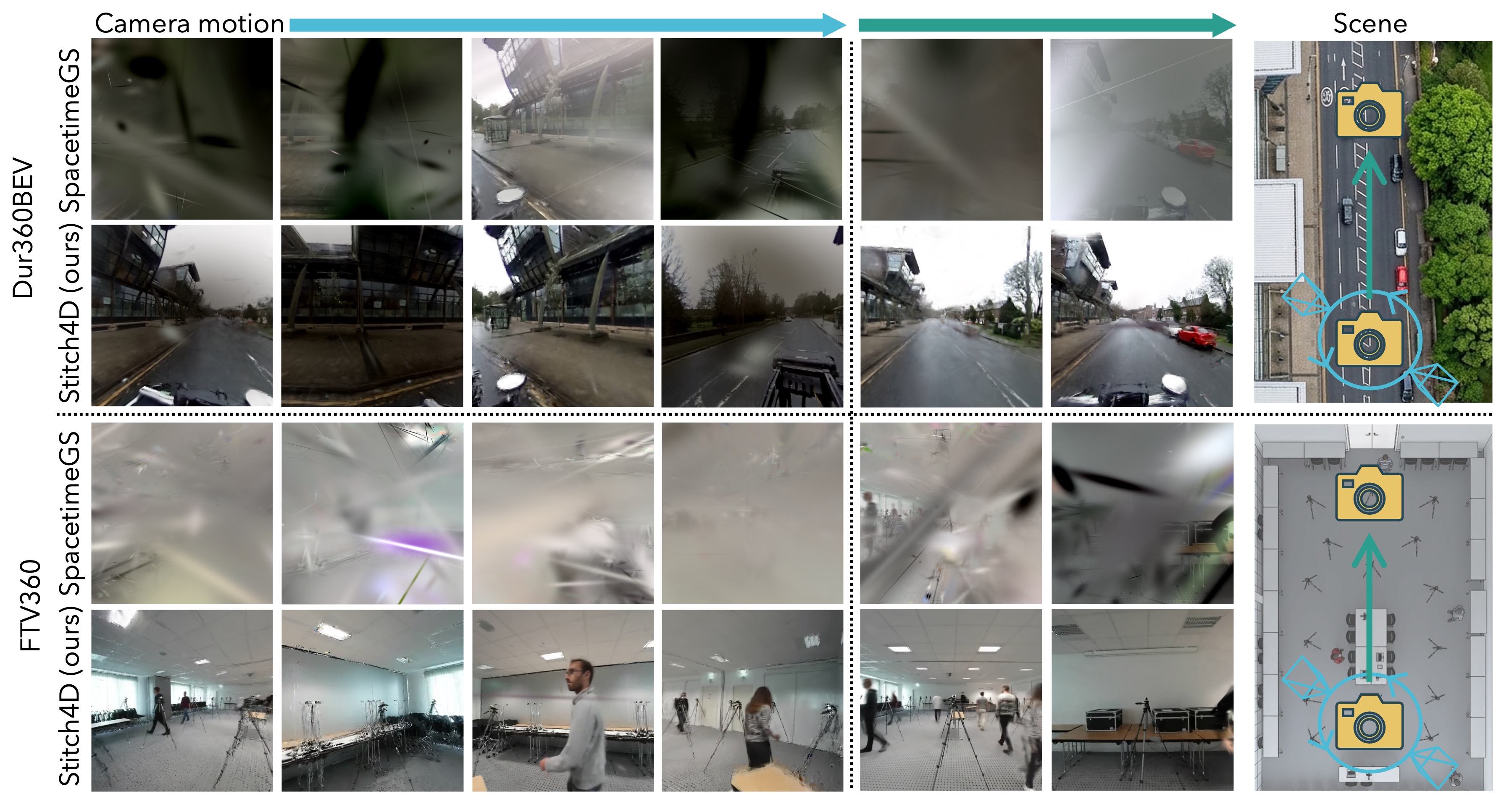}
    \vspace{-8mm}
    \caption{Additional real-world qualitative results on Dur360BEV~\cite{dur360bev} and FTV360~\cite{maugey2019ftv360}.}
    \vspace{-4.7mm}
    \label{fig:real_supp}
\end{figure*}

Fig.~\ref{fig:real_supp} shows additional qualitative results on real-world scenes.
These results further demonstrate that Stitch4D can reconstruct dynamic urban scenes from sparse multi-location panoramic videos beyond the synthetic benchmark.
Compared with the baseline methods, Stitch4D produces more coherent scene geometry and sharper appearance across viewpoints, especially in regions between input camera locations.

\subsubsection{Trajectory Interpolation and Seen-viewpoints.}
\begin{figure*}[t]
    \centering
    \includegraphics[height=0.43\textheight]{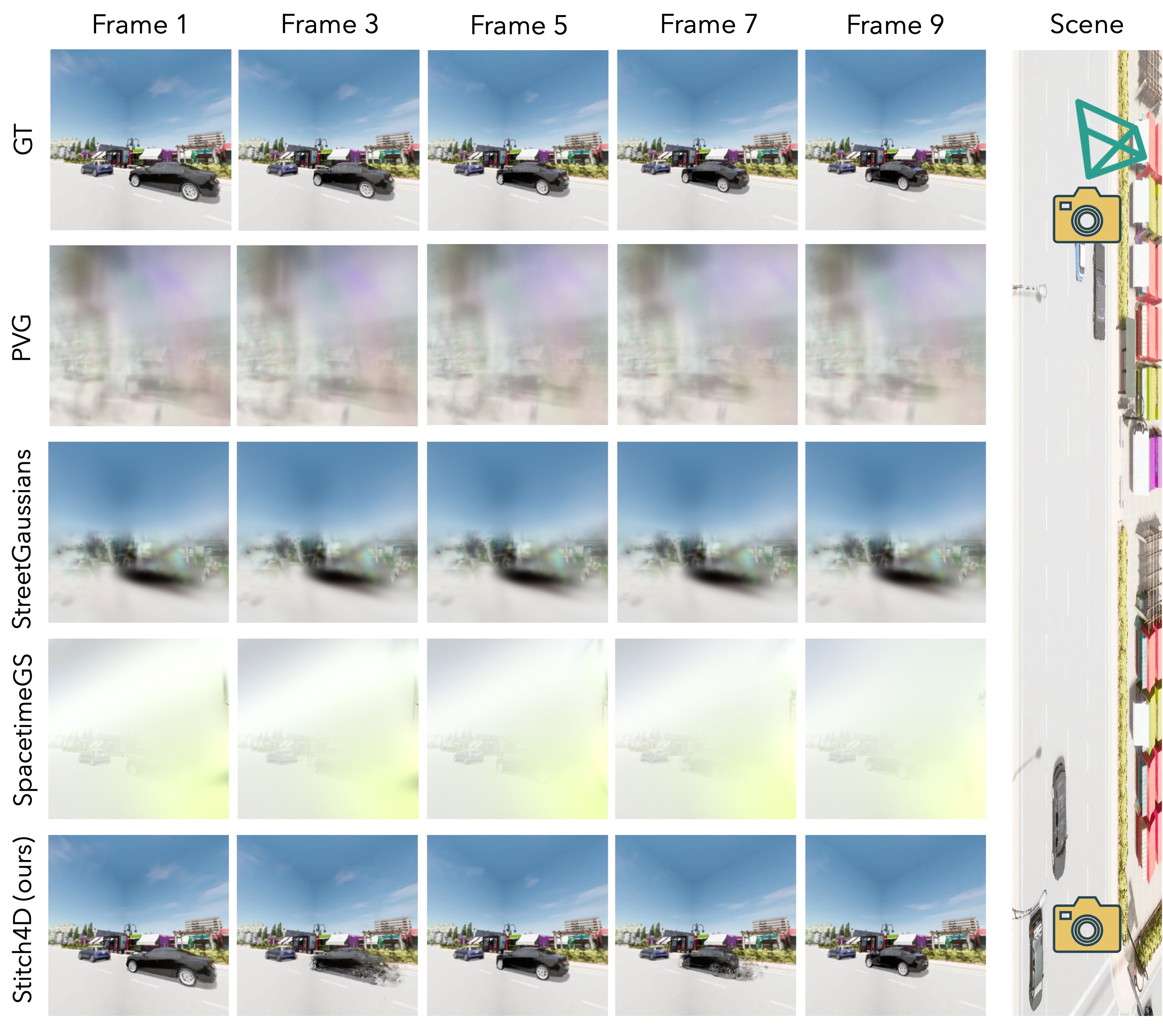}
    \caption{
    Qualitative results in the temporal split setting (seen-viewpoints condition) for Urban Area 3. 
    }
    \label{fig:qual_temporal_virtual_2}
    \vspace{-2mm}
\end{figure*}
\begin{figure*}[t]
    \centering
    \includegraphics[width=0.8\linewidth]{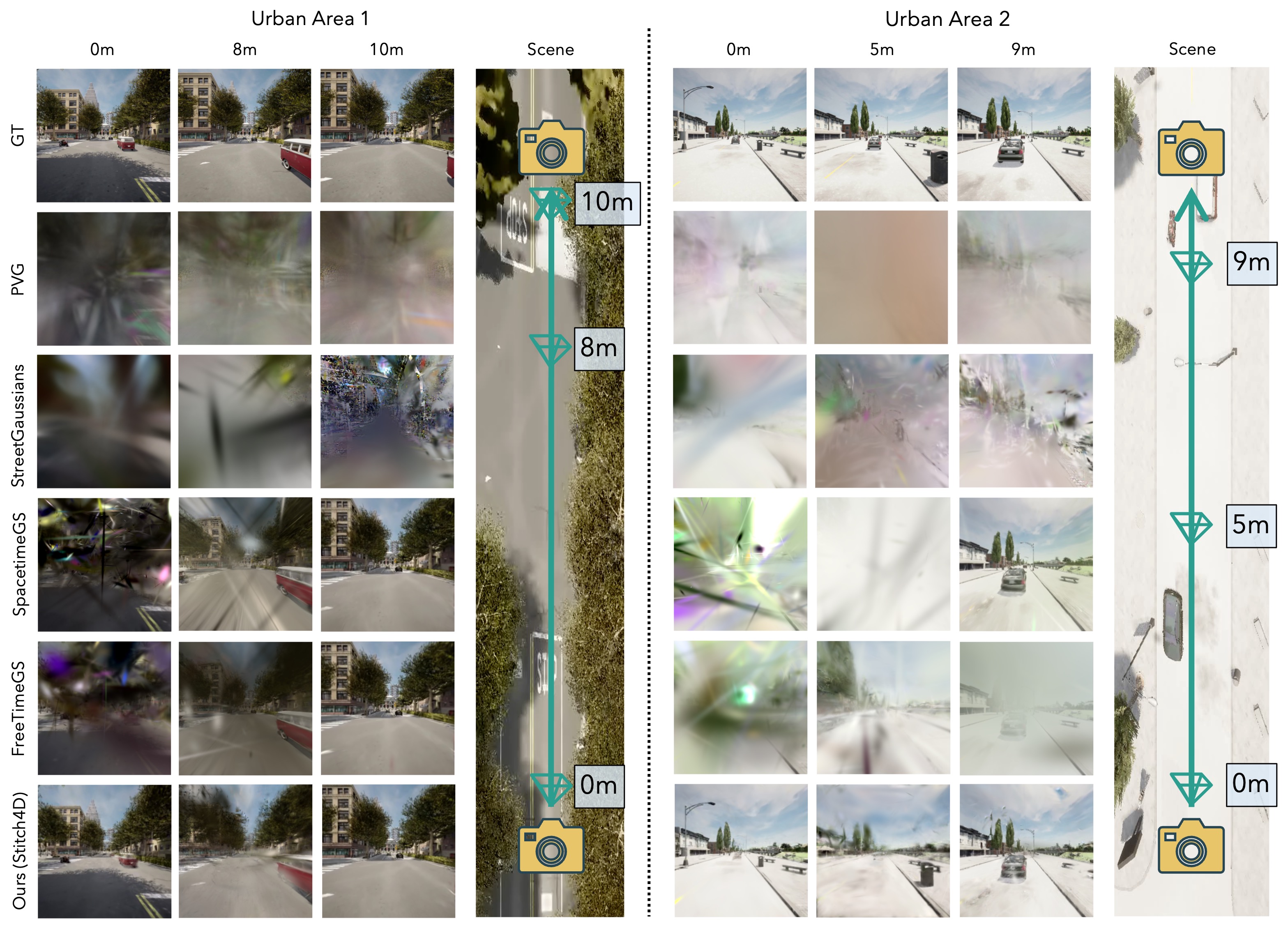}
    \caption{
    Qualitative results in the full reconstruction setting (trajectory interpolation condition). 
    Yellow cameras indicate input viewpoints, while green icons denote evaluation viewpoints along the trajectory. 
    }
    \label{fig:inter}
    \vspace{-3mm}
\end{figure*}
\begin{figure*}[t]
    \centering
    \includegraphics[width=\linewidth]{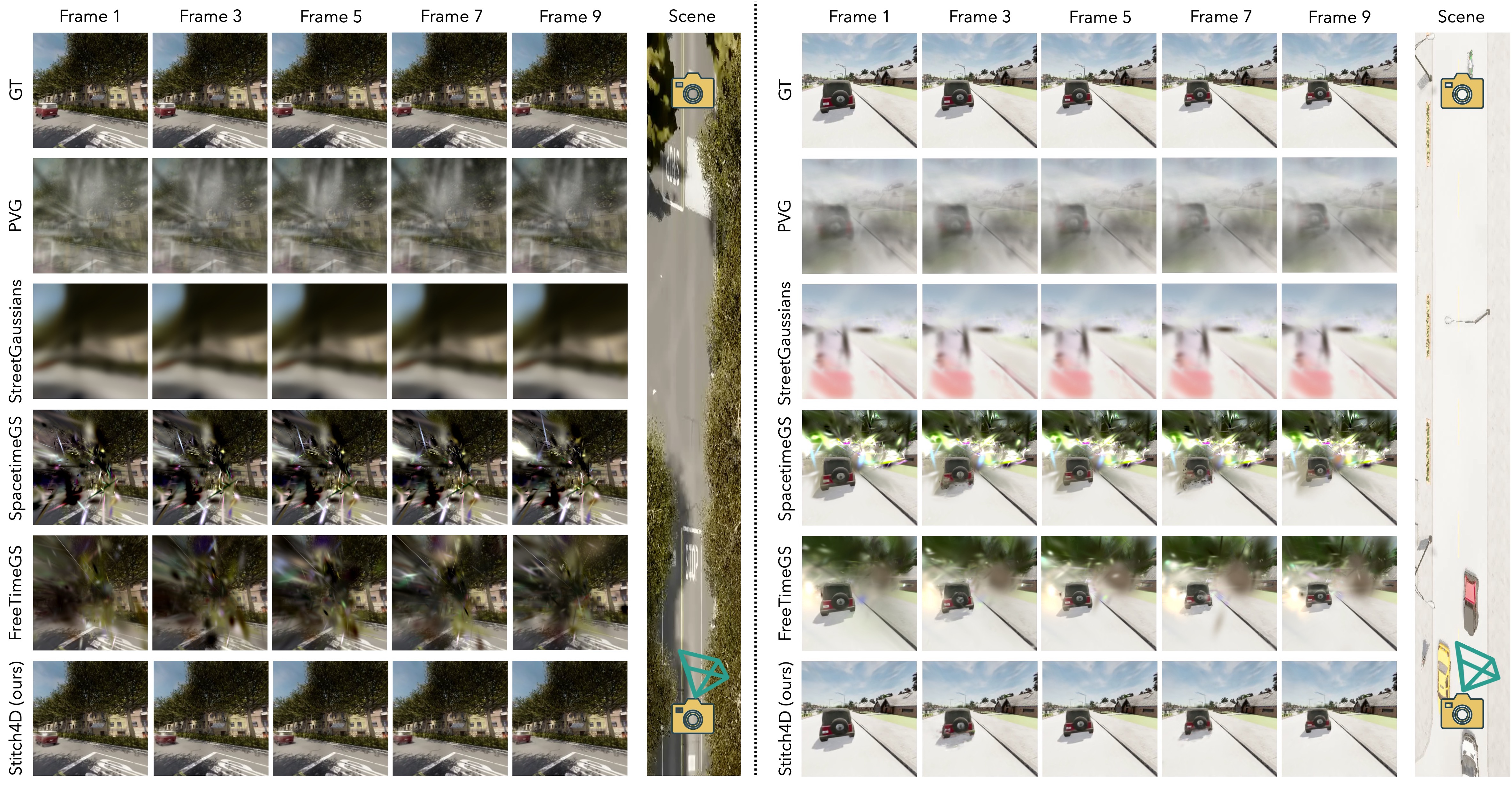}
    \caption{
    Qualitative results in the temporal split setting (seen-viewpoints condition) for Urban Area 1 (left) and Urban Area 2 (right). 
    Each row corresponds to a fixed virtual camera and each column denotes the frame index (test frames: 3 and 7). 
    Camera icons indicate the camera locations of the input videos, while green icons represent the rendering viewpoints.
    }
    \label{fig:seen_viewpoints}
    \vspace{-3mm}
\end{figure*}

Figs.~\ref{fig:qual_temporal_virtual_2}, ~\ref{fig:inter}, and ~\ref{fig:seen_viewpoints} compare the qualitative results of the proposed method with the baselines.
Fig.~\ref{fig:qual_temporal_virtual_2} provides additional seen-viewpoint results for Urban Area~3 under the temporal split setting.
Fig.~\ref{fig:inter} shows the trajectory interpolation condition in the full reconstruction setting, whereas Fig.~\ref{fig:seen_viewpoints} shows the seen-viewpoints condition in the temporal split setting.
In Fig.~\ref{fig:inter}, each column renders the scene from the virtual viewpoint indicated at the top.
In Fig.~\ref{fig:seen_viewpoints}, each row corresponds to a fixed virtual camera, while each column corresponds to the frame index shown at the top (test frames: 3 and 7).

\begin{figure*}[t]
    \centering
    \includegraphics[width=\linewidth]{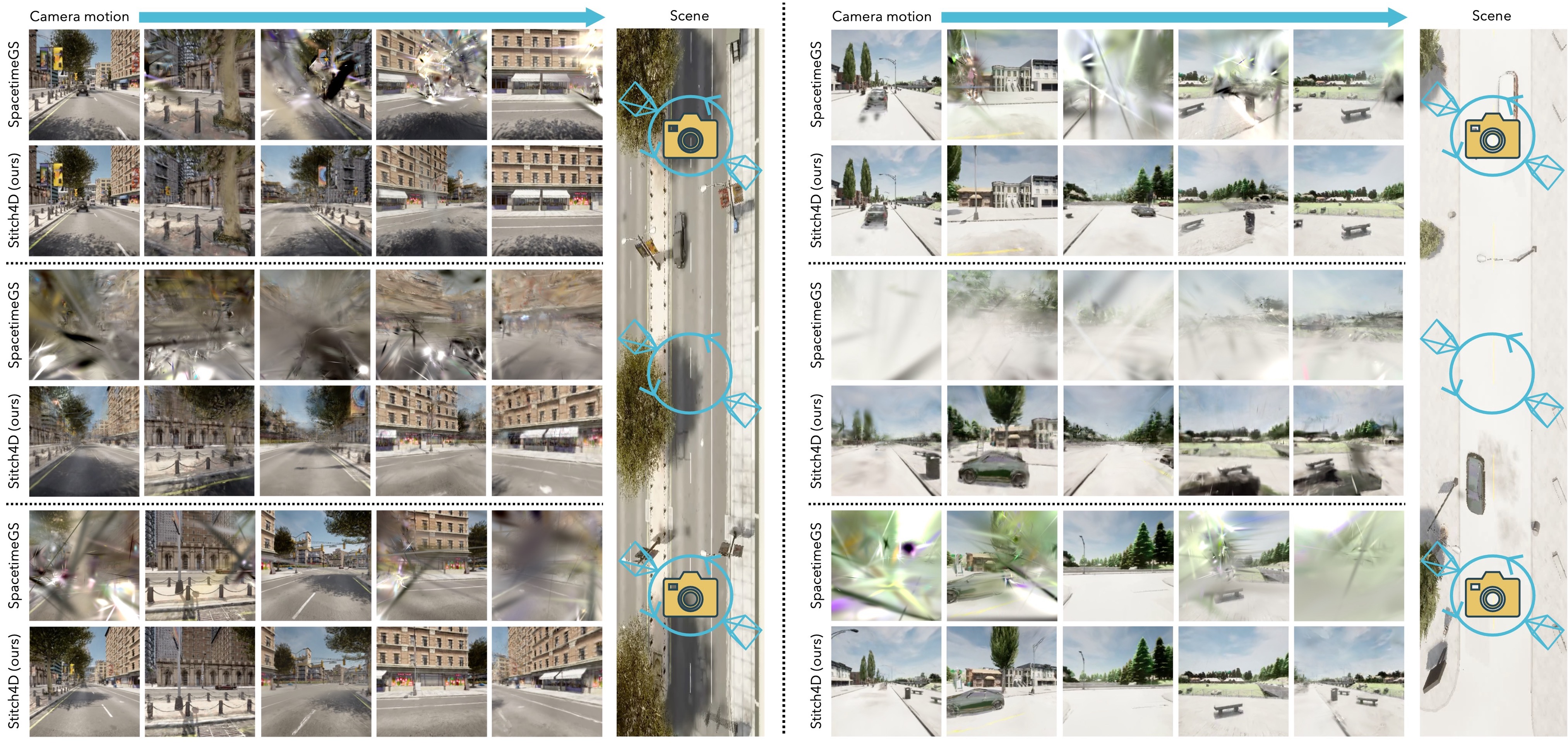}
    \caption{
    Qualitative comparison on Urban Area 1 (left) and Urban Area 2 (right) under the full reconstruction setting with freely moving camera trajectories.
    Each row compares SpacetimeGS and our method (Stitch4D).
    Each column represents frames rendered along the camera motion using the rotateshow trajectory.
    The camera trajectory used for rendering is illustrated on the right.
    Camera icons denote the camera locations of the input videos.
    }
    \label{fig:rotateshow}
\end{figure*}
\begin{figure*}[t]
    \centering
    \includegraphics[width=\linewidth]{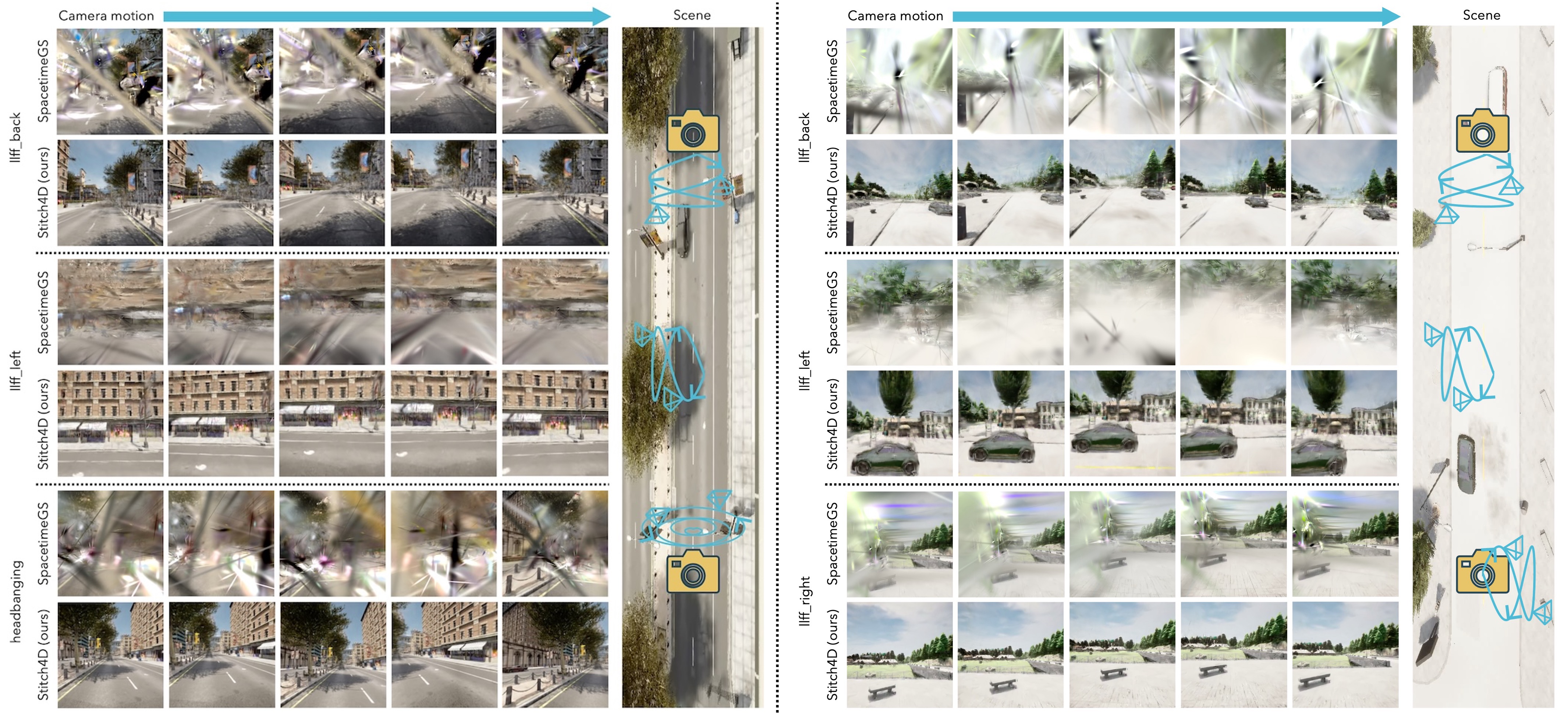}
    \caption{
    Qualitative comparison on Urban Area 1 (left) and Urban Area 2 (right) under the full reconstruction setting with freely moving camera trajectories.
    }
    \label{fig:trajectory}
\end{figure*}
\noindent\textbf{Freely Moving Trajectories.}
Figs.~\ref{fig:rotateshow} and 
\ref{fig:trajectory} compare the qualitative results of Stitch4D with the best-performing baseline, SpacetimeGS, which achieves the highest PSNR among the baselines.
All results are rendered under the full reconstruction setting with freely moving camera trajectories.
The rendering trajectories are illustrated on the right side of each figure.
Fig.~\ref{fig:rotateshow} presents results rendered with the \textbf{rotateshow} trajectory.
In contrast, 
Fig.~\ref{fig:trajectory} evaluates rendering under additional trajectories, including LLFF-style directional variants (\textbf{llff\_back}, \textbf{llff\_left}, \textbf{llff\_right}) and the \textbf{headbanging} trajectory.
We consider three types of camera trajectories.
First, the \textbf{rotateshow} trajectory rotates the camera around the scene while maintaining limited translational motion.
Second, the \textbf{llff\_back}, \textbf{llff\_left}, and \textbf{llff\_right} trajectories move the camera along a smooth local orbit around the scene while observing it from different directions, corresponding to yaw offsets of 180$^\circ$, -90$^\circ$, and +90$^\circ$, respectively.
Third, the \textbf{headbanging} trajectory introduces large oscillatory rotations of the camera viewpoint.

\noindent\textbf{Scaling to More Input Videos.}
\begin{figure*}[!th]
    \centering
    \includegraphics[height=0.44\textheight]{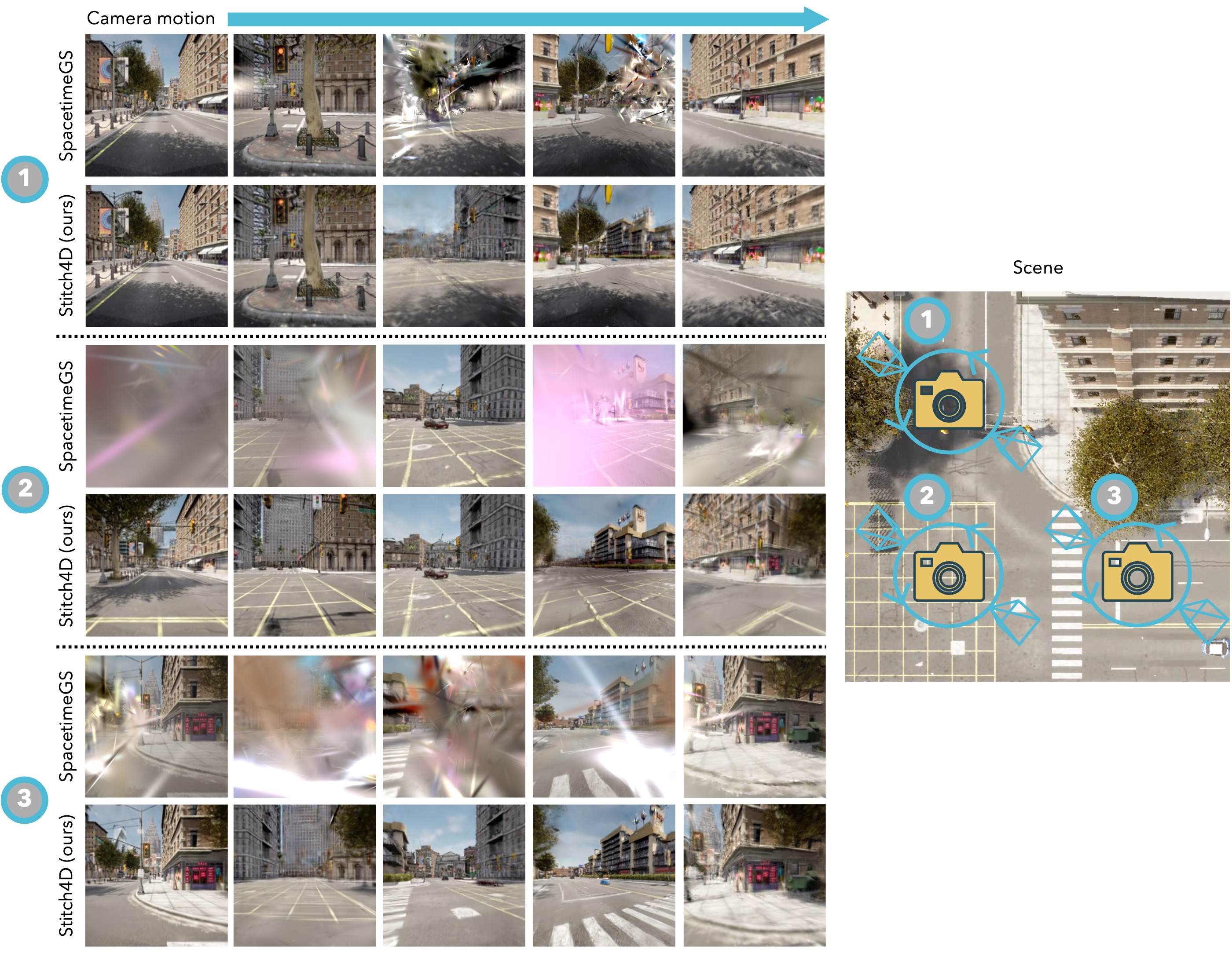}
    \caption{
    Additional qualitative comparison on Urban Area 1 under the full reconstruction setting using the rotateshow camera trajectory. The scene is reconstructed from three input videos.
    }
    \vspace{-2mm}
    \label{fig:3input_rotateshow}
\end{figure*}

We further evaluate the method with an increased number of input videos.
Fig.~\ref{fig:3input_rotateshow} presents qualitative results obtained using three input videos.


\noindent\textbf{Failure Case.}
\begin{figure*}[!th]
    \centering
    \includegraphics[height=0.43\textheight]{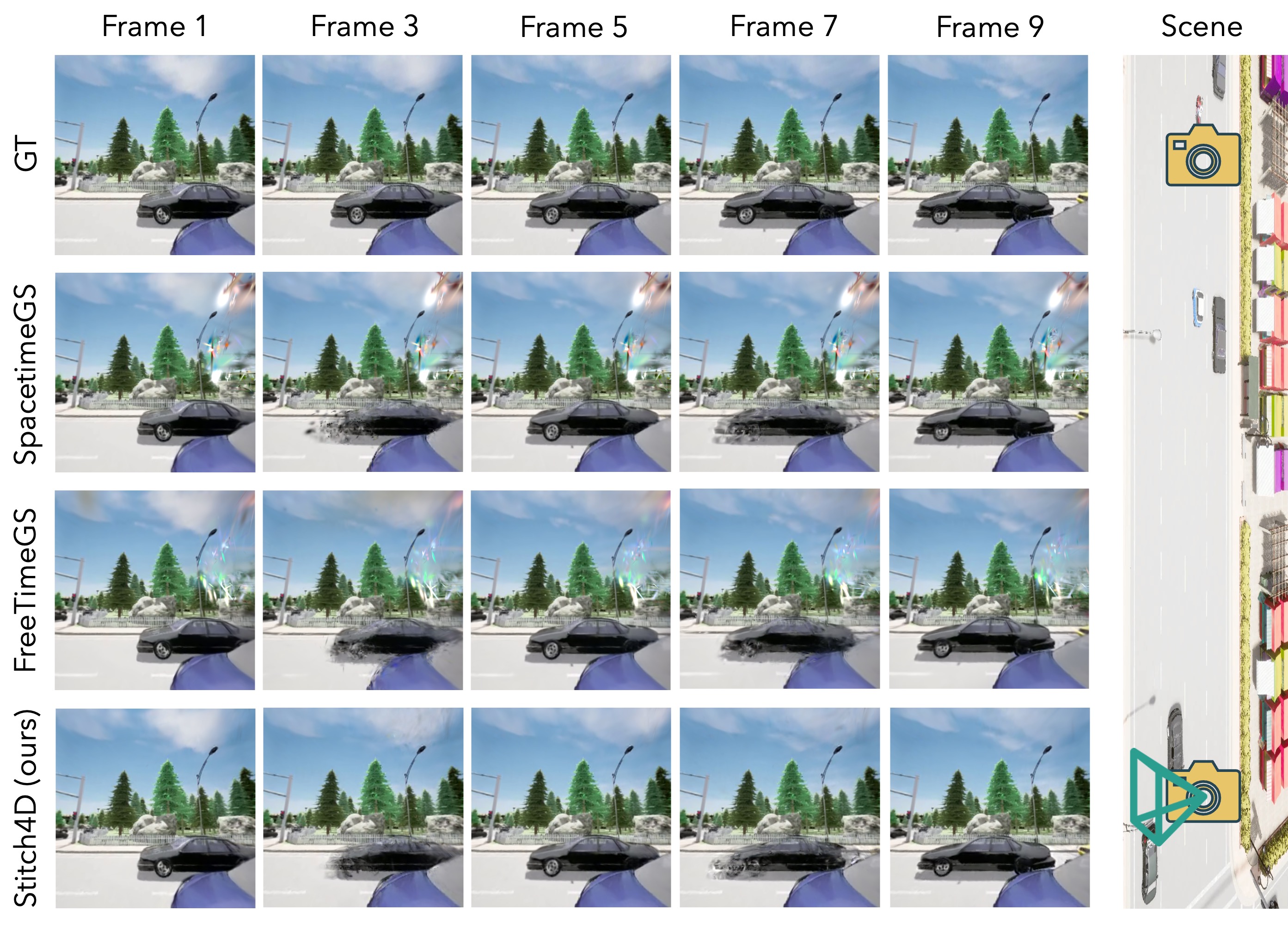}
    \caption{
    Qualitative results illustrating a failure case in the temporal split setting (seen-viewpoints condition) for Urban Area 3. 
    }
    \label{fig:fail}
    \vspace{-3mm}
\end{figure*}
Fig.~\ref{fig:fail} presents a representative failure case comparing Stitch4D with the baseline methods, SpacetimeGS and FreeTimeGS, under the seen-viewpoints condition in the temporal split setting.
Each row corresponds to a fixed virtual camera, while each column corresponds to the frame index shown at the top (test frames: 3 and 7).
In this example, the dynamic object (the moving car) exhibits noticeable distortions at the test frames, where the object boundaries collapse and the shape becomes temporally inconsistent.
This suggests that reconstructing dynamic objects remains challenging under sparse inter-location observations.


%
%

\end{document}